\documentclass[journal]{IEEEtran}

\usepackage{times}
\usepackage{epsfig}
\usepackage{graphicx}
\usepackage{amsmath}
\usepackage{amssymb}

\usepackage{caption}
\usepackage{subcaption}
\usepackage{epsfig}
\usepackage{amsmath}
\usepackage{amssymb}
\usepackage{booktabs}
\usepackage{rotating}
\usepackage{algorithm} 
\usepackage{algpseudocode}
\usepackage{multirow}
\usepackage{color}

\usepackage{enumitem}
\usepackage{enumerate}
\definecolor{dgreen}{rgb}{0.412,0.741,0.271}
\definecolor{dblue}{rgb}{0.220,0.325,0.639}
\definecolor{dred}{rgb}{0.933,0.122,0.137}
\usepackage{mathtools}

\DeclarePairedDelimiter\floor{\lfloor}{\rfloor}

\newcommand{\etal}{\textit{et al}.}

\newcommand{\eg}{\textit{e}.\textit{g}.}

\newif\ifshowcomments
\showcommentstrue

\ifshowcomments
\newcommand{\TODO}[1]{{\color{red}{[TODO: #1]}}}
\newcommand{\revised}[1]{{\color[rgb]{0.2,0.2,0.7}{#1}}}
\newcommand{\lzhu}[1]{{\color[rgb]{0.7,0.7,0}{#1}}}
\newcommand{\phil}[1]{{\color[rgb]{0.9,0.1,0.1}{#1}}}

\else
\newcommand{\TODO}[1]{}
\newcommand{\revised}[1]{}
\newcommand{\lzhu}[1]{}
\newcommand{\phil}[1]{}
\fi

\ifCLASSINFOpdf
\else
\fi
\usepackage[pagebackref=false,breaklinks=true,letterpaper=true,colorlinks,citecolor=black,linkcolor=black,bookmarks=false]{hyperref}

\begin{document}
	%
	\title{SAC-Net: Spatial Attenuation Context \\ for Salient Object Detection}
	%
	%
	%
	
	\author{Xiaowei~Hu, 
		Chi-Wing~Fu, 
		Lei~Zhu,
		Tianyu Wang, 
		and~Pheng-Ann~Heng
		\IEEEcompsocitemizethanks{
			\IEEEcompsocthanksitem X. Hu, L. Zhu, and T. Wang are with the Department of Computer Science and Engineering, The Chinese University of Hong Kong, Hong Kong SAR, China.
			\IEEEcompsocthanksitem C.-W. Fu and P.-A. Heng are with the Department of Computer Science and Engineering, The Chinese University of Hong Kong, Hong Kong SAR, China and also with Shenzhen Key Laboratory of Virtual Reality and Human Interaction Technology, Shenzhen Institutes of Advanced Technology, Chinese Academy of Sciences, Shenzhen 518055, China.
			%
		}
	}

	\maketitle
	
	\begin{abstract}
		This paper presents a new deep neural network design for salient object detection by maximizing the integration of local and global image context within, around, and beyond the salient objects.
		Our key idea is to adaptively propagate and aggregate the image context features with variable attenuation over the entire feature maps.
		To achieve this, we design the spatial attenuation context (SAC) module to recurrently translate and aggregate the context features independently with different attenuation factors and then to attentively learn the weights to adaptively integrate the aggregated context features.
		By further embedding the module to process individual layers in a deep network, namely SAC-Net, we can train the network end-to-end and optimize the context features for detecting salient objects.
		Compared with 29 state-of-the-art methods, experimental results show that our method performs favorably over all the others on six common benchmark data, both quantitatively and visually.
	\end{abstract}
	

	%
	\IEEEpeerreviewmaketitle



\section{Introduction}
\label{sec:introduction}

Salient object detection aims to distinguish the most visually distinctive objects from an input image and it is an effective pre-processing step in many image processing and computer vision tasks, \eg, object segmentation~\cite{wang2015saliency} and tracking~\cite{hong2015online}, video compression~\cite{hadizadeh2014saliency} and abstraction~\cite{zhao2009real}, image editing~\cite{cheng2010repfinder}, texture smoothing~\cite{zhu2018saliency}, as well as few-shot learning~\cite{zhang2019few}.
It is a fundamental problem in computer vision research and has been extensively studied in the past decade.

Early works attempt to detect salient objects based on low-level cues like contrast, color, and texture~\cite{cheng2015global,jiang2013salient,liu2011learning,deng2019saliency}.
However, relying on low-level cues is clearly inadequate to finding salient objects, which involve high-level semantics.
%
Hence, most recent methods~\cite{li2016deepsaliency,luonon2017,wang2016saliency,zhang2017learning,zhao2015saliency} employ convolutional neural networks (CNNs) and take a data-driven approach to the problem by leveraging both high-level semantics and low-level details extracted from multiple CNN layers~\cite{chen2018reverse,deng18r,hou2017deeply,Hu_2018_AAAI,li2016deep,li2018contour,zhang2018bi,zhang2017amulet,zhang2018progressive,zhang2019salient,wang2019salient,zhao2019pyramid,wu2019cascaded,liu2019simple,wang2019iterative,wu2019mutual,li2018contrast}.
%
%
However, since the convolution operator in CNN processes a local neighborhood in the spatial domain~\cite{wang2018non}, existing methods tend to miss global spatial semantics in the results,~\eg, they may
misrecognize background noise as salient objects; see Section~\ref{subsec:compare_state_of_the_art} for quantitative and qualitative comparisons.

%
%

\begin{figure} [tp]
	\centering
	\includegraphics[width=0.99\linewidth]{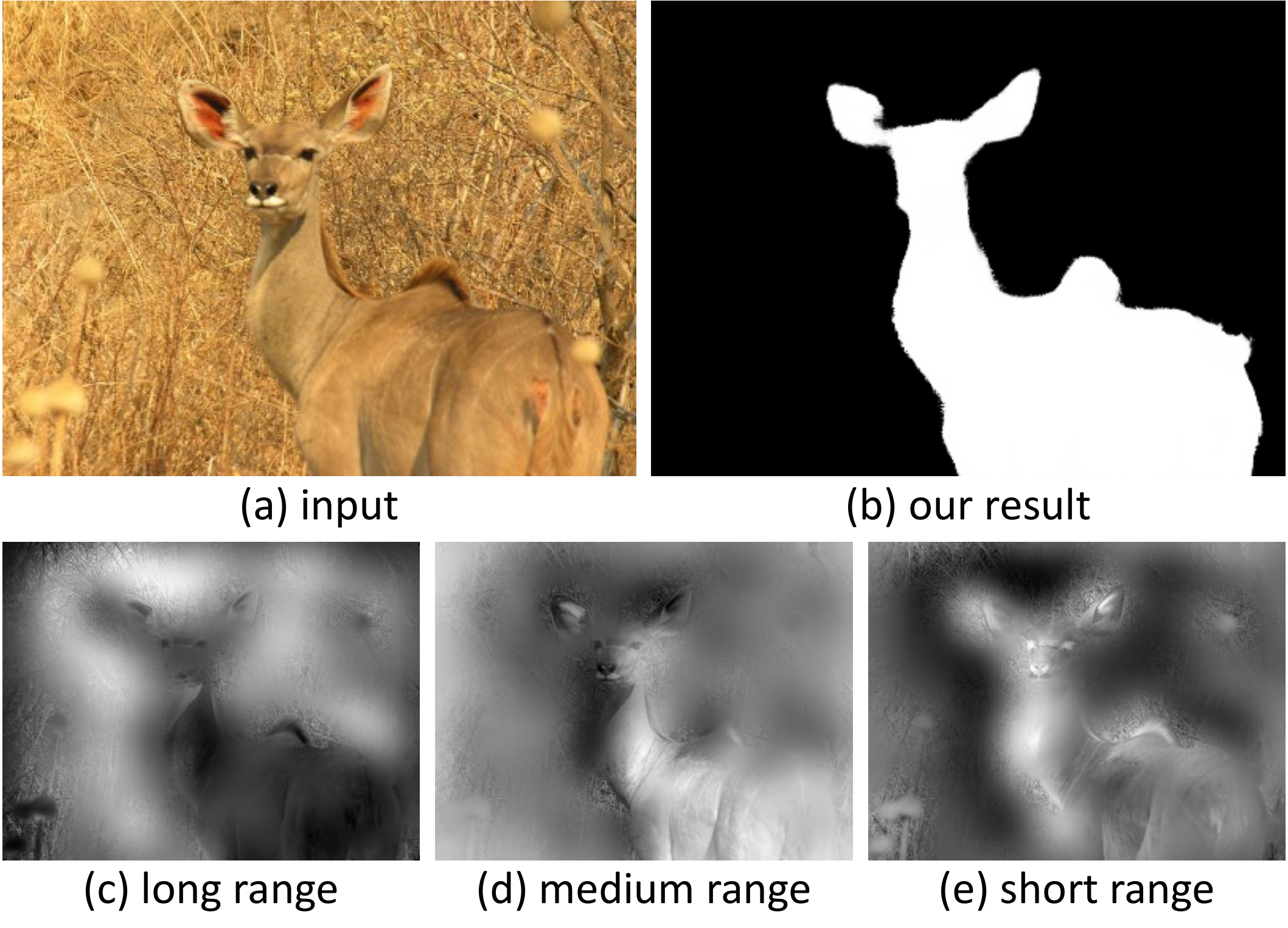}

	\caption{A challenging example (a), where our method is still able to find the object contour (b); see (c)-(e) for the attention weights learned for the context with different spatial ranges.}
		
%
%
	\label{fig:motivation}

\end{figure}

Essentially, salient objects are key elements that stand out from the background.
%
Such an inference process~\cite{goferman2012context} should involve not only the local image context within and around the salient objects, but also the global image context, as well as a suitable integration of the various context features.
Ideally, after extracting context features per image pixel, if we can connect all these features and let them communicate with every other over the spatial domain, we can optimize the feature integration for maximized performance.
However, it is computationally infeasible in practice.
Hence, we present to propagate context features with different attenuation factors over the spatial domain of the image and learn to aggregate the resulting features adaptively; by then, our network can learn to detect salient objects by adaptively considering context features within, around, and even far from, the salient objects.

%

To achieve this, we present the spatial attenuation context, which is achieved by the following steps: (i) the image context is aggregated by propagating the information pixel by pixel over the whole feature maps; as a result, each pixel will obtain the global information from all other pixels of the feature maps; (ii) the propagation ability is affected by an attenuation factor, where a large attenuation factor reduces the information propagation and leads to a short-range context while a small attenuation factor improves the information propagation and leads to a long-range context; (iii) the image contexts with different ranges are dynamically merged by learning a set of attention weights.
Fig.~\ref{fig:motivation} shows a challenging example with the associated attention maps learned in our network for {\em integrating the various image context\/}:
(c) long-range context aggregated with a small attenuation factor
helps locate the global background;
(d) medium-range context helps identify the image regions of the same object; and
(e) short-range context aggregated with a large attenuation factor helps locate the boundary between salient and non-salient regions.
Please see the supplementary material for detailed explanations with more illustrations.

In details, we formulate the {\em spatial attenuation context module\/}, or {\em SAC module\/} for short, in a deep network to allow the image features in a CNN to {\em propagate over variable spatial ranges by articulating different attenuation factors in the propagation\/}.
Our module has two rounds of recurrent translations to propagate and aggregate the image features.
In each round, we propagate features independently using different attenuation factors towards different directions in the spatial domain; further, we formulate an attention mechanism to learn the weights to combine the aggregated features.
Hence, we can adopt different attenuation factors (or influence ranges) for different image features.
Furthermore, we deploy an SAC module in each layer of our network and predict a saliency map per layer based on the output from the SAC module and the convolutional features.
Below, we summarize the major contributions of this work:
\begin{itemize}[]
\item
We design the spatial attenuation context (SAC) module to recurrently propagate the image features over the whole feature maps with variable attenuation factors and learn to adaptively integrate the features through an attention mechanism in the module.
Then, we adopt the SAC module in each layer of our network architecture to learn the spatial attenuation context in different layers, and train the whole network in an end-to-end manner for salient object detection.

\item
We evaluate our method and compare it against 29 state-of-the-art methods on six common benchmark data.
Results show that our method performs favorably over all the others for all the benchmark data.
Our code, trained models, and predicted saliency maps are publicly available at \url{https://xw-hu.github.io/}.

%
%
\end{itemize}


\section{Related Work}
\label{sec:related}

%

Rather than being comprehensive, we discuss mainly the methods on single-image salient object detection.
Early methods use hand-crafted priors such as image contrast~\cite{jiang2013salient,perazzi2012saliency}, color~\cite{borji2012exploiting,mahadevan2013biologically}, texture~\cite{yan2013hierarchical,yang2013saliency}, and other low-level visual cues~\cite{harel2007graph}; see~\cite{borji2015salient} for a survey.
Clearly, hand-crafted features are insufficient to capture high-level semantics, so methods based on them often fail for nontrivial inputs.

Recent works~\cite{li2016deepsaliency,luonon2017,zhao2015saliency}
exploit convolutional neural networks (CNN) to learn deep features for detecting salient objects.
%
However, since these methods just take features at deep CNN layers, they tend to miss the details in the salient objects, which are captured mainly in the shallow layers.
Several recent works~\cite{wang2016saliency,zhang2017learning,li2016deep,hou2017deeply,zhang2017amulet,Hu_2018_AAAI,deng18r,zhang2018bi,zhang2018progressive,li2018contour,chen2018reverse,wang2018detect,zhang2019salient} enhance the detection quality by further integrating features in multiple CNN layers to simultaneously leverage more global and local context in the inference process.
Among them,
Li~\etal~\cite{li2016deep} explored the semantic properties and visual contrast of salient objects,
Hou~\etal~\cite{hou2017deeply} created short connections to integrate features in different layers, while
Zhang~\etal~\cite{zhang2017amulet} derived a resolution-based feature combination module and a boundary-preserving refinement strategy.
Hu~\etal~\cite{Hu_2018_AAAI} recurrently aggregated deep features to exploit the complementary saliency information between the multi-level features and the features at each individual layer.
Later, Deng~\etal~\cite{deng18r} adopted residual learning to alternatively refine features at deep and shallow layers.
Zhang~\etal~\cite{zhang2018bi} formulated a bi-directional message passing model to select features for integration.
Zhang~\etal~\cite{zhang2018progressive} designed an attention-guided network to progressively select and integrate multi-level information.
Li~\etal~\cite{li2018contour} used a two-branch network to simultaneously predict the contours and saliency maps.
Chen~\etal~\cite{chen2018reverse} leveraged residual learning and reverse attention to refine the saliency maps.
Li~\etal~\cite{li2018contrast} presented a contrast-oriented deep neural network, which adopts two network streams for both dense and sparse saliency inference.
Zhang~\etal~\cite{zhang2019salient} designed a symmetrical CNN to learn the complementary saliency information and presented a weighted structural loss to enhance the boundaries of salient objects.
Wang~\etal~\cite{wang2018detect} explored the global and local spatial relations in deep networks to locate salient objects and refine the object boundary.
%
Although the detection quality keeps improving, the exploration of global spatial context, particularly in the shallow layers,
is still heavily limited by the convolution operator in CNN, which is essentially a local spatial filter~\cite{wang2018non}.

\begin{figure*} [htbp]
	\centering
	\includegraphics[width=0.9\linewidth]{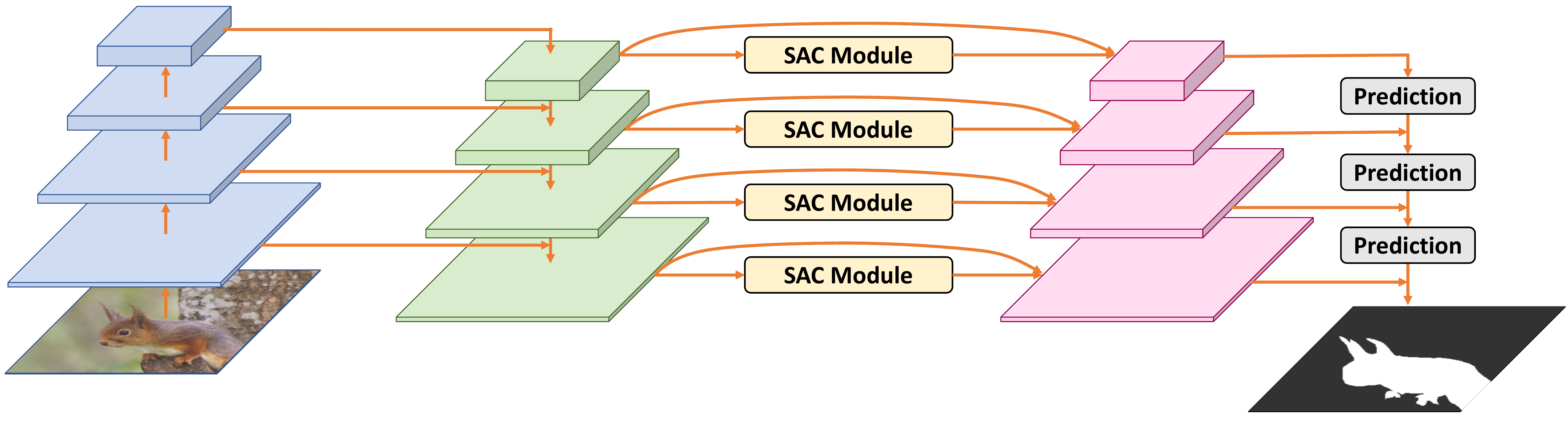}
	\caption{The schematic illustration of our spatial attenuation context network (SAC-Net):
		(i) extract feature maps (in blue) in different resolutions from the input image using a convolutional neural network;
		(ii) construct a feature pyramid (in green) by successively upsampling the feature map at a deep layer and combining the upsampled result with the feature map at an adjacent shallower layer;
		(iii) use SAC modules (see Fig.~\ref{fig:context}) to generate spatial attenuation context features for each layer;
		(iv) concatenate the outputs from the SAC modules with the convolutional features (in red);
		and
		(v) lastly, successively predict a saliency map at each layer and take the final saliency map of the largest resolution as the network output.
		In the figure, feature maps are indicated by blocks and thicker blocks of smaller sizes are higher-level features at deeper layers.}
	\label{fig:arc}
\end{figure*}

Very recently,
Liu and Han~\cite{liu2018deep} incorporated global context and scene context by developing a deep spatial long short-term memory model.
Liu~\etal~\cite{liu2018picanet} aggregated the attended contextual features from a global/local view in feature maps of varying resolutions.
Wang~\etal~\cite{wang2019salient} presented a pyramid attention structure and leveraged the salient edge information to better segment salient objects.
Feng~\etal~\cite{feng2019attentive} designed an attentive feedback network to further explore the boundaries of the salient objects.
Zhao and Wu~\cite{zhao2019pyramid} used the dilated convolution and channel-wise \& spatial attention to aggregate multi-scale context features.
Wu~\etal~\cite{wu2019cascaded} proposed to discard the feature maps at shallow layers for acceleration and used the saliency map generated from one network branch to refine the features of another branch.
Liu~\etal~\cite{liu2019simple} introduced two pooling-based modules to progressively refine the highly semantic features for detail enriched saliency maps.
Wang~\etal~\cite{wang2019iterative} predicted the saliency maps by iteratively aggregating the feature maps in the top-down and bottom-up manner.
Zhang~\etal~\cite{zhang2019capsal} incorporated the semantic information of salient objects from the image captions.
Qin~\etal~\cite{qin2019basnet} formulated a boundary-aware salient object detection network by combining a deeply supervised encoder-decoder and a residual refinement module, and leveraged a hybrid loss to optimize the whole network.
Wu~\etal~\cite{wu2019mutual} jointly performed foreground contour detection and edge detection tasks by using multi-task intertwined supervision.
Fu~\etal~\cite{fu2019deepside} presented a Deepside to incorporate hierarchical CNN features and fused multiple side outputs based on a segmentation-based pooling.
Li~\etal~\cite{li2017instance} developed a multiscale saliency refinement network, which is used for instance-level salient object segmentation.
Zhu~\etal~\cite{zhu2019aggregating} learned the attentional dilated features to detect the salient objects.
Even the detection performance continues to improve on the benchmarks~\cite{li2015visual,li2014secrets,martin2001database,wang2017learning,yan2013hierarchical,yang2013saliency},
current methods may still miss local parts in salient objects and misrecognize noises in non-salient regions as salient objects.
Except the above works,
	Song~\etal~\cite{song2019multi} presented a novel multi-scale attention network for accurate object detection.
	Peng~\etal~\cite{peng2019two} proposed two-stream collaborative learning with a spatial-temporal attention approach for video classification.
	He~\etal~\cite{he2019and} developed a multi-scale and multi-granularity deep reinforcement learning approach for fine-grained visual categorization.
	Peng~\etal~\cite{peng2018object} formulated an object-part attention model for weakly supervised fine-grained image classification.

The recent works~\cite{liu2018deep,liu2018picanet} that emphasize the importance of reasoning spatial context for salient object detection.
	Comparing with the PiCANet~\cite{liu2018deep,liu2018picanet}, which aggregates the global context formation on the feature maps with small resolutions by adopting the expensive long short-term memory models and aggregates the local context information on the feature maps with large resolutions through convolutions,
	%
	we leverage and selectively aggregate surrounding image context spatially in the same CNN layer by a new concept, i.e., spatial attenuation context, which attentively allows the context features to recurrently translate with varying attenuation factors (including local and global information) on the feature maps with any resolutions.
%





\section{Methodology}
\label{sec:method}

Fig.~\ref{fig:arc} outlines the architecture of our spatial attenuation context network (SAC-Net), which takes a whole image as input and predicts the saliency map in an end-to-end manner.
First, we use a CNN to generate feature maps in different resolutions and progressively propagate the image features at deep layers to feature maps at shallow layers to construct a feature pyramid~\cite{lin2017feature}.
%
After that, we use our SAC modules to harvest spatial attenuation context per layer and concatenate the module outputs with the corresponding convolutional features.
%
Lastly, we predict a result per layer, upsample and merge it with the shallower-layer output, and take the result of the largest resolution as the final network output.
In the following subsections, we first elaborate on the SAC module, and then present the strategies to train and test our network for salient object detection.

\begin{figure} [tp]
	\centering
	\includegraphics[width=0.98\linewidth]{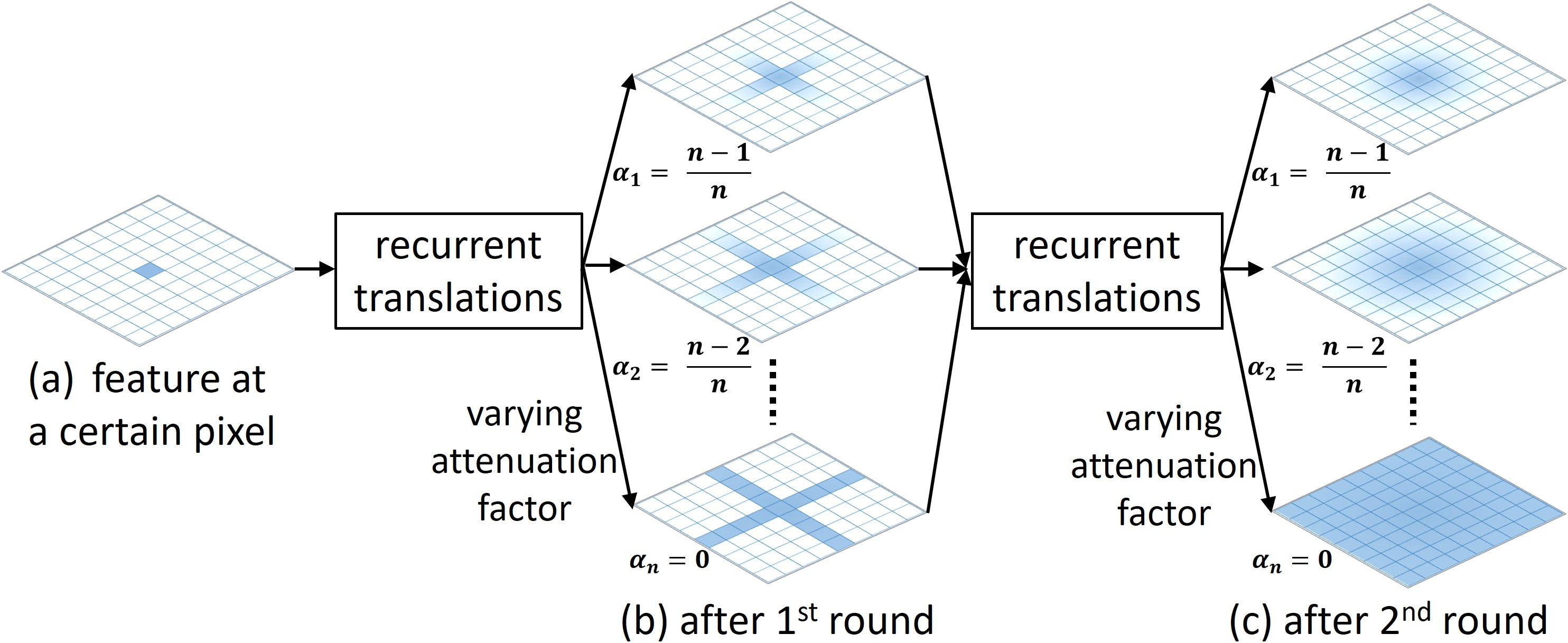}
	\caption{Illustrating how the image features propagate with varying attenuation factors ($\alpha_k$) inside the SAC module; please see Fig.~\ref{fig:context} for the detailed module architecture.}
	\label{fig:information_flow}
\end{figure}

\begin{figure*} [tp]
	\centering
	\includegraphics[width=1\linewidth]{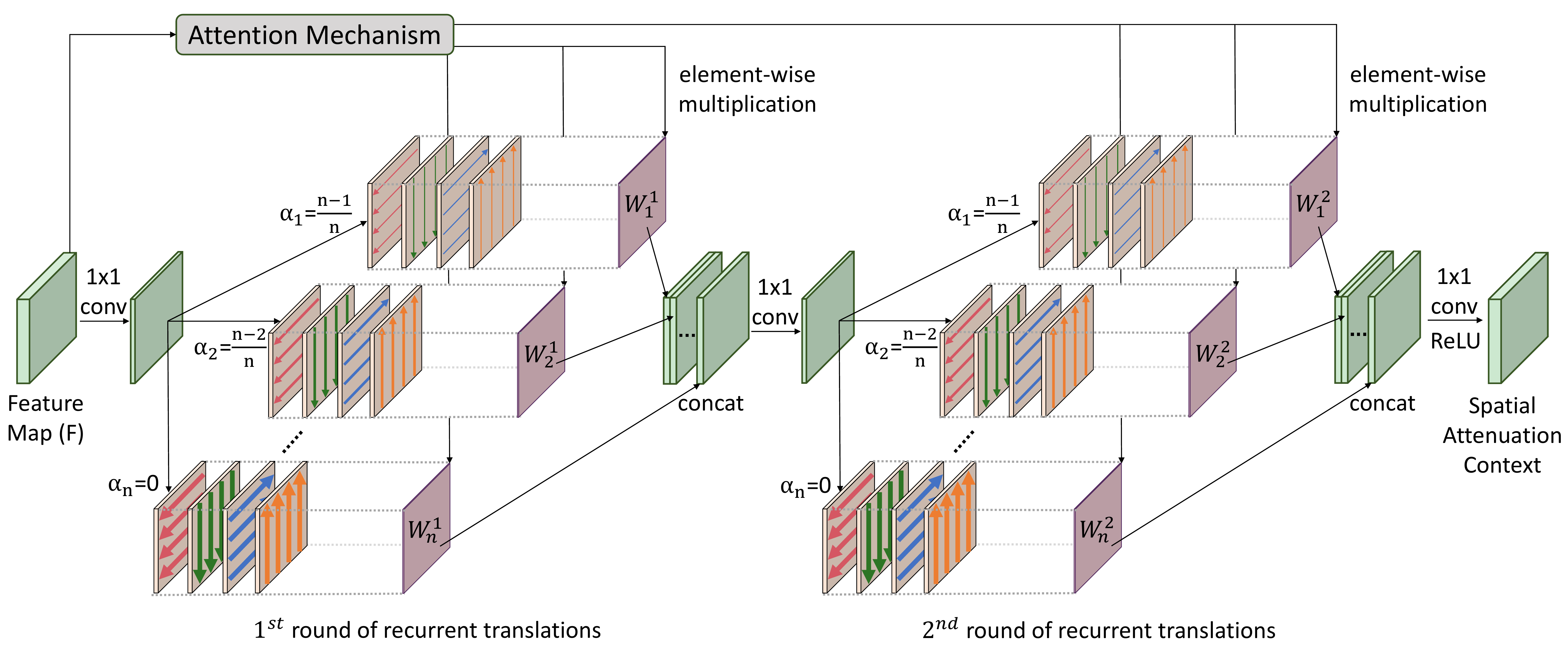}
	\caption{The schematic illustration of the spatial attenuation context (SAC) module.
		We adopt two rounds of recurrent translations to propagate and aggregate image features.
		In each round, the colored arrows show the recurrent translation direction, while thicker (or thinner) arrows indicate stronger (or weaker) information propagation with less (or more) attenuation.}
	\label{fig:context}
	\vspace*{-1mm}
\end{figure*}

\subsection{Spatial Attenuation Context Module}
\label{subsec:SACmodule}

Fig.~\ref{fig:context} shows the architecture of the {\em spatial attenuation context module\/}, or {\em SAC module\/}, which takes a feature map as input and produces spatial attenuation context in the same resolution.
As presented earlier,
the spatial attenuation context contains image context aggregated by propagating local image context using varying attenuation factors via an attention mechanism; hence, we can disperse the local image context adaptively over the whole feature maps.



See again the SAC module in Fig.~\ref{fig:context}.
First, we use a $1\times1$ convolution on the input feature map to reduce the number of feature channels.
Then, we adopt recurrent translations with varying attenuation factors ($\alpha_k$) to disperse the local image features in four different directions; see the illustration in Fig.~\ref{fig:information_flow}(b) \& the detailed structure of recurrent translations in Fig.~\ref{fig:context}.
At this moment, each pixel learns the spatial attenuation context along the four directions.
After two rounds of recurrent translations, we adaptively disperse the local features over the 2D domain; see Fig.~\ref{fig:information_flow}(c).
Hence, each pixel knows \emph{the global spatial attenuation context over the entire feature map}.
More importantly, we learn the weights to combine the recurrently-aggregated results via an attention mechanism in an end-to-end manner (Fig.~\ref{fig:context}), so each pixel in the SAC module output can receive spatial context \emph{adaptively} from its surroundings;
please see the supplementary material for the detail explanations.



\textbf{Recurrently-attenuating translation.}
To optimize the dispersal of local context,
we first formulate a parametric model to recurrently aggregate the image features with attenuation.
Given the feature map after a $1$~$\times$~$1$ convolution (see Fig.~\ref{fig:context}), we recurrently translate its features using different attenuation factors $\alpha_k$ in four principal directions: left, up, right, and down.
Moreover, to ensure manageable memory consumption, we set the number of feature channels in each recurrently-aggregated feature map as $\floor*{\frac{256}{n}}$, where $n$ is the number of different attenuation factors in the SAC module; see Table~\ref{table:architecture_analysis} for an experiment on $n$.


Denoting $f_{i,j}$ as the feature at pixel $(i, j)$ in a feature map, our recurrently-attenuating translation process propagates features progressively over the spatial domain using the following equation (typically in the up direction) :
\begin{eqnarray}
\label{eq:propagation}
f_{i,j}^{up}(\alpha_k,\beta)
&=&
\max\big(r^{up}_{i,j}, 0 \big) + \beta \min \big(r^{up}_{i,j}, 0 \big)
\nonumber
\\
\text{and} \ \
r^{up}_{i,j}
&=&
(1-\alpha_k) \cdot f_{i-1,j}^{up}+f_{i,j} \ ,
\end{eqnarray}
where
$\alpha_k =\frac{n-k}{n}$ ($k \in \{1,2,...,n\}$)
is the attenuation factor,
and $\beta$ is a learnable parameter in our recurrently-attenuating translation model.

In Eq.~\eqref{eq:propagation}, we recurrently aggregate image features by using $r^{up}_{i,j}$, where a smaller $\alpha_k$ (close to zero) allows the features to propagate over a longer distance, while a larger $\alpha_k$ (close to one) limits the propagation, so the related local features affect a smaller local area; see again the illustration in Fig.~\ref{fig:information_flow}.
Moreover, when $r^{up}_{i,j} < 0$, the first term in $f_{i,j}^{up}$ will become zero, and $\beta$ will be multiplied with $r^{up}_{i,j}$.
We define $\beta$ in Eq.~\eqref{eq:propagation} to reduce the feature magnitude when it is negative.
Since we learn the value of $\beta$ for each feature channel, we can introduce nonlinearities when aggregating the spatial context and express more complex relations among the local features.
Note that in our experiments, we initialize $\beta$ as $0.1$ for all the feature channels and learn it automatically during the network training process; in practice, we found that $\beta$ rarely goes beyond one in our experiments.


\begin{table*}  [htbp]
	\begin{center}
		\setlength\tabcolsep{3pt}
		\caption{Comparing our method (SAC-Net) with 29 state-of-the-art methods using the $F_\beta$, $S_m$ and MAE metrics.
Top two results are highlighted in {\color{dred}\bf red} and {\color{dblue}\bf blue}, respectively; ``-'' indicates results that are not publicly available on the corresponding dataset; and ``*'' indicates CRF is used as a post-processing step in the methods.}
		\label{table:state-of-the-art}
		\resizebox{1.0\textwidth}{!}{%
			\begin{tabular}{|c|c|ccc|ccc|ccc|ccc|ccc|ccc|}
				\hline
				Dataset &	-&
				\multicolumn{3}{c|}{ECSSD~\cite{yan2013hierarchical}} &
				\multicolumn{3}{c|}{PASCAL-S~\cite{li2014secrets}} &
				\multicolumn{3}{c|}{SOD~\cite{movahedi2010design}} &
				\multicolumn{3}{c|}{HKU-IS~\cite{li2015visual}} &
				\multicolumn{3}{c|}{DUT-OMRON~\cite{yang2013saliency}} &
				\multicolumn{3}{c|}{DUTS-test~\cite{wang2017learning}} 
	
				\\
				
				\cline{1-20}
				 Metric & Year & $F_\beta$ & $S_m$ & MAE & $F_\beta$ & $S_m$ & MAE & $F_\beta$ & $S_m$ & MAE & $F_\beta$ & $S_m$ & MAE & $F_\beta$ & $S_m$ & MAE & $F_\beta$ & $S_m$ & MAE \\
				\hline

				\textbf{SAC-Net* (ours)} & - & {\color{dred}\bf 0.954}& {\color{dred}\bf 0.930} & {\color{dred}\bf 0.028} 
				& {\color{dred}\bf 0.876} & {\color{dred}\bf 0.801}& {\color{dred}\bf 0.070} 
				& {\color{dred}\bf 0.884} & {\color{dred}\bf 0.801}& {\color{dred}\bf 0.092} 
				&{\color{dred}\bf 0.945} & {\color{dred}\bf 0.925} & {\color{dred}\bf 0.023} 
				&	{\color{dred}\bf 0.832} & {\color{dred}\bf 0.846} &  {\color{dred}\bf 0.050} 
				& {\color{dred}\bf 0.898} & {\color{dred}\bf 0.878}& {\color{dred}\bf 0.032} 
				\\

				PiCA-RC*~\cite{liu2018picanet} &2018 & {\color{dblue}\bf 0.940} & {\color{dblue}\bf 0.916} & {\color{dblue}\bf 0.035} &
				{\color{dblue}\bf 0.870} & {\color{dblue}\bf 0.789} & {\color{dblue}\bf 0.073} & 
				{\color{dblue}\bf 0.867} & {\color{dblue}\bf 0.780} &  {\color{dblue}\bf 0.094} & 
				{\color{dblue}\bf 0.929} & {\color{dblue}\bf 0.905}&  {\color{dblue}\bf 0.031} & 
				{\color{dblue}\bf 0.828} & {\color{dblue}\bf 0.826} & {\color{dblue}\bf 0.054} & 
				{\color{dblue}\bf 0.871} & {\color{dblue}\bf 0.849} & {\color{dblue}\bf 0.040}  \\

				R$^3$Net*~\cite{deng18r} &2018 &  0.935 & 0.910 & 0.040 
				& 0.845 & 0.749&  0.100 
				& 0.847 & 0.761&  0.124
				& 0.916 &  0.900& 0.036
				& 0.805 & 0.817& 0.063 
				&0.833 & 0.823 &0.058 \\
				
				GNLB*~\cite{zhu2018saliency} &2018 & 0.931 & 0.900 & 0.045
				& 0.840 & 0.758 & 0.096
				& 0.837 & 0.744 & 0.127
				& 0.917 & 0.886 & 0.037
				& 0.800 & 0.817 & 0.058
				& 0.830 & 0.811 & 0.058 \\

				RADF*~\cite{Hu_2018_AAAI} &2018 & 0.924& 0.894 & 0.049 
				& 0.832& 0.754 & 0.102
				& 0.835 &0.759 & 0.125 
				& 0.914 & 0.889& 0.039 
				& 0.789 & 0.815& 0.060 
				& 0.819 & 0.814 & 0.061 \\

				DSS*~\cite{hou2017deeply} &2017 &0.916 & 0.882 & 0.053
				& 0.829& 0.739& 0.102
				& 0.842& 0.746 & 0.118 
				& 0.911& 0.881 & 0.040 
				& 0.771& 0.790 & 0.066 
				& 0.825& 0.812 & 0.057 \\

				DCL*~\cite{li2016deep} &2016& 0.898&0.868& 0.071
				& 0.822& 0.783& 0.108
				& 0.832& 0.745 & 0.126
				& 0.904& 0.861& 0.049 
				& 0.757& 0.771& 0.080 
				&0.782& 0.795& 0.088 \\
				
				\hline
				
				\textbf{SAC-Net (ours)} &- & {\color{dred}\bf 0.951} & {\color{dred}\bf0.931} & {\color{dred}\bf 0.031} & 
				{\color{dred}\bf 0.879}& {\color{dred}\bf0.806} & {\color{dred}\bf 0.070} & 
				{\color{dred}\bf 0.882} & {\color{dred}\bf0.809} & {\color{dred}\bf 0.093} & 
				{\color{dred}\bf 0.942} & {\color{dred}\bf0.925} & {\color{dred}\bf 0.026}  & 
				{\color{dred}\bf 0.830} & {\color{dred}\bf0.849} & {\color{dred}\bf 0.052} & 
				{\color{dred}\bf 0.895} & {\color{dred}\bf0.883} & {\color{dred}\bf 0.034} \\

				PoolNet-R~\cite{liu2019simple} &2019 
				&{\color{dblue}\bf 0.944}& {\color{dblue}\bf 0.921} & 0.039 
				&0.865& 0.794& 0.080 
				&{\color{dblue}\bf 0.869} & {\color{dblue}\bf 0.801} & {\color{dblue}\bf 0.100} 
				&{\color{dblue}\bf 0.934} & {\color{dblue}\bf 0.912} & 0.033 
				& {\color{dred}\bf  0.830} & {\color{dblue}\bf 0.836} & {\color{dblue}\bf 0.056} 
				&  {\color{dblue}\bf 0.886} &  {\color{dblue}\bf 0.871} &  {\color{dblue}\bf 0.040} \\
				
				BASNet~\cite{qin2019basnet} & 2019 
				& 0.942&0.916& {\color{dblue}\bf 0.037} 
				&0.858&0.785&0.084 
				&0.851&0.772&0.112 
				&0.929&0.909& {\color{dblue}\bf 0.032}  
				&0.811& {\color{dblue}\bf 0.836} & {\color{dblue}\bf 0.056} 
				&0.860&0.853&0.047 \\

				CPD-R~\cite{wu2019cascaded} &2019 
				& 0.939 & 0.918 & 0.037 
				& 0.861& 0.789 & 0.078 
				&  0.859 & 0.771 & 0.110 
				& 0.925 & 0.906 & 0.034 
				& 0.797 & 0.825 & {\color{dblue}\bf 0.056} 
				& 0.865 & 0.858 & 0.043 \\
				
				AFNet~\cite{feng2019attentive} &2019 
				& 0.935 & 0.917 & 0.042 
				& 0.866 & 0.792 & 0.076 
				&-&-&- 
				&0.925&0.905& 0.036
				& {\color{dblue}\bf 0.820}& 0.826 & 0.057 
				& 0.867 & 0.855 & 0.045 \\

				MLMSNet~\cite{wu2019mutual} &2019 
				&0.930&0.909&0.045 
				&0.858&0.790&0.079 
				&0.862&0.790&0.106 
				&0.922& 0.906 &0.039 
				&0.793&0.809&0.064 
				&0.854&0.851&0.048 \\

				CapSal~\cite{zhang2019capsal} & 2019 
				&-&-&- 
				&0.868 &0.769&0.079 
				&-&-&- 
				&0.889&0.849&0.057 
				&-&-&- 
				&0.845&0.808&0.060 \\
				
				
				PiCA-R~\cite{liu2018picanet} &2018 
				& 0.935 & 0.917&  0.047 & 
				0.868 &  {\color{dblue}\bf 0.800} &  0.078 &  
				0.864 & 0.793&  0.103 &  
			    0.919 & 0.904 &  0.043 &  
				0.820&  0.832 &  0.065 
				& 0.863 &   0.859 &  0.050  \\
				
				ASNet~\cite{wang2018salient}  &2018 & 0.932& 0.915 & 0.047
				& {\color{dblue}\bf 0.869}& 0.794 & 0.075
				& 0.859 & 0.800 & 0.105
				& 0.922 & 0.906 & 0.041
				& -&- &-
				&0.835 &0.834&0.060\\

				R$^3$Net~\cite{deng18r} &2018 & 0.929 & 0.910&  0.051 
				& 0.842 & 0.761& 0.103 
				& 0.839 & 0.770& 0.131 
				& 0.914 & 0.897&  0.046 
				&  0.802 & 0.819  & 0.073 
				&0.831 & 0.829 & 0.067 \\
		
				BDMPM~\cite{zhang2018bi}  &2018 & 0.928& - &  0.044
				&  0.862 & - &  {\color{dblue}\bf 0.074}
				& 0.851 & - & 0.106 
				& 0.920 &-& 0.038 
				& - & - & - 
				& 0.850 &-& 0.049 \\
		
				
				PAGRN~\cite{zhang2018progressive} &2018 & 0.927& 0.889 &0.061
				 &0.849 & 0.749 & 0.094
				  &- &- & - 
				  &0.918 &0.887& 0.048 
				  & 0.771& 0.775 & 0.071 
				  & 0.854 & 0.825 &0.055 \\

				GNLB~\cite{zhu2018saliency} &2018& 0.926 & 0.904 & 0.056 
				& 0.841 & 0.772 & 0.099
				& 0.834 & 0.762 & 0.133
				& 0.909 & 0.891 & 0.048
				& 0.800 & 0.824 & 0.067
				& 0.821 & 0.822 & 0.068 \\
			
				DGRL~\cite{wang2018detect} &2018 &0.925& 0.906 &0.045 
				& 0.850 & 0.796& 0.080 
				& 0.846 & 0.777& 0.104 
				& 0.914 & 0.897 & 0.037 
				& 0.779& 0.810 &0.063
				& 0.834 & 0.836 & 0.051 \\

				RAS~\cite{chen2018reverse} &2018
				& 0.916 & 0.893 & 0.058 
				& 0.842 & 0.735&  0.122 
				& 0.847 & 0.767& 0.123 
				& 0.913 & 0.887 & 0.045 
				& 0.785 & 0.814 &  0.063
				&0.831 & 0.828 & 0.059 \\
			
				C2S~\cite{li2018contour} &2018
				& 0.911 & 0.896 &0.053 
				& 0.845 & 0.793& 0.084 
				& 0.821 & 0.763&  0.122 
				& 0.898 & 0.889 &0.046 
				&0.759 & 0.799&0.072 
				&0.811 & 0.822 & 0.062  \\

				SRM~\cite{wang2017stagewise}  &2017 & 0.917 & 0.895 & 0.054 
				& 0.847& 0.782 & 0.085
				 & 0.839 & 0.746& 0.126 
				 & 0.906 & 0.888 & 0.046 
				 & 0.769& 0.798& 0.069 
				 & 0.827 & 0.825 & 0.059 \\

				Amulet~\cite{zhang2017amulet} &2017 & 0.913 & 0.894 & 0.059
				& 0.828 & 0.794 & 0.095 
				&0.801& 0.755 & 0.146
				& 0.887& 0.886 &0.053 
				& 0.737& 0.781& 0.083 
				& 0.778& 0.796 & 0.085 \\
		
				UCF~\cite{zhang2017learning}  &2017 & 0.910& 0.883 &0.078
				& 0.821& 0.792& 0.120
				& 0.800& 0.763& 0.164
				& 0.886& 0.875 & 0.073
				 & 0.735& 0.758& 0.131
				  &0.771 & 0.777& 0.117 \\
				  
				NLDF~\cite{luonon2017} &2017 &0.905 & 0.875 &0.063
				& 0.831 & 0.756& 0.099 
				&0.810& 0.759&  0.143
				& 0.902& 0.879 & 0.048 
				& 0.753 & 0.770& 0.080 
				& 0.812& 0.815 & 0.066  \\
	
				DHSNet~\cite{liu2016dhsnet}  &2016 &0.907& 0.884 & 0.059
				& 0.827& 0.752 & 0.096
				& 0.823& 0.752 & 0.127
				& 0.892& 0.870 & 0.052 
				& -&-&- 
				& 0.807& 0.811 &  0.067\\

				RFCN~\cite{wang2016saliency} &2016 & 0.898 & 0.860 & 0.097
				& 0.827& 0.793&  0.118
				& 0.805& 0.717& 0.161
				& 0.895& 0.859&  0.079 
				& 0.747 & 0.774 & 0.095 
				& 0.784& 0.791 & 0.091 \\
			
				ELD~\cite{lee2016deep} &2016 &0.867& 0.841 &0.080
				& 0.771& -& 0.121
				& 0.760& -& 0.154
				& 0.844& -& 0.071 
				& 0.719& 0.751 & 0.091 
				& 0.738& 0.719 & 0.093 \\
	
				MDF~\cite{li2015visual} &2015 & 0.831& 0.764 & 0.108
				& 0.759& 0.692 &0.142
				& 0.785& 0.674 &0.155
				& -& - & - 
				& 0.694 & 0.703 & 0.092 
				& 0.730 & 0.723 & 0.094 \\
	
				LEGS~\cite{wang2015deep} &2015 & 0.827& 0.787  & 0.118
				& 0.756& 0.682 &0.157
				& 0.707& 0.661 &0.215
				& 0.770& - &0.118 
				& 0.669 & - & 0.133 
				& 0.655& - & 0.138 \\
			
				BSCA~\cite{qin2015saliency} &2015 & 0.758& 0.725& 0.183
				& 0.666& 0.633 &0.224
				& 0.634& 0.622 & 0.266
				& 0.723& 0.700 &0.174 
				& 0.616 & 0.652& 0.191 
				& 0.597& 0.630 & 0.197  \\
				
				DRFI~\cite{jiang2013salient} &2013 & 0.786& -& 0.164
				& 0.698& -&0.207
				& 0.697& -&0.223
				& 0.777& -&0.145 
				& - & - &- 
				& 0.647& -& 0.175 \\

				\hline
				SAC-Net (Res50) & - & 0.945 & 0.924 & 0.034 
				& 0.871 & 0.805 & 0.072 
				& 0.872 & 0.804 & 0.093 
				& 0.936 & 0.920 & 0.028
				& 0.808 & 0.832 & 0.057 
				& 0.881 & 0.873 & 0.037
				\\
				\hline

		\end{tabular} }
	\end{center}
\vspace{-2mm}
\end{table*}

\textbf{Attention mechanism.}
\label{sec:att_mechanism}
After recurrently-translating the input feature map using different attenuation factors in four directions, we will obtain $4n$ feature maps; see the feature maps with colored arrows in Fig.~\ref{fig:context}.
%
As discussed earlier, the long-range image context reveals global semantics, while the short-range context helps identify the boundary between salient and non-salient regions.
To adaptively leverage the complementary advantages of these aggregated spatial context features, we formulate an attention mechanism to learn the weights for selectively integrating them.

As shown at the top left corner in Fig.~\ref{fig:context}, we take the input feature map $F$ as the input to the attention mechanism and produce a set of unnormalized attention weights $\{ A_1^1, A_2^1, ..., A_n^1 \}$, each corresponding to a particular attenuation factor; superscript $1$ indicates that these weights are for the first round of recurrent translations.
Then, we apply the Softmax function (Eq.~\eqref{normalize_attention}) to normalize the weights and produce the attention weight maps $\{W_{1}^1, W_{2}^1, ..., W_{n}^1\}$ associated with different attenuation factors (see Fig.~\ref{fig:context}):
%
%
%
\begin{eqnarray}
\label{attention}
\{ A_1^1, A_2^1, ..., A_n^1 \}
& = &
\Psi \ ( F; \ \theta ) \ , \ \text{and}
\\
\label{normalize_attention}
w_{i,j,k}^1
& = &
\frac{exp \ (a_{i,j,k}^1)}{\sum_{k} \ exp \ (a_{i,j,k}^1)} \ ,
\end{eqnarray}
where
$a_{i,j,k}^1 \in A_{k}^1$ is the unnormalized attention weight at pixel $(i,j)$ for attenuation factor $\alpha_k$,
$w_{i,j,k}^1 \in W_{k}^1$ are the normalized attention weights, and
$\theta$ denotes the parameters learned by $\Psi$, which consists of two $3\times3$ convolution layers and one $1\times1$ convolution layer, and we apply the group normalization~\cite{wu2018group} and ReLU non-linear operation~\cite{krizhevsky2012imagenet} after the first two convolution layers.

Next, we multiply $W_{k}^1$ with the corresponding context features aggregated after the recurrent translations:
\begin{eqnarray}
\label{equ:four_direction_concat_att}
\hspace*{-3mm}
f_{i,j}
&
\hspace*{-2.5mm}
=
\hspace*{-2.5mm}
&
\oplus_{k=1}^{n} \ \Big[ \ \big( \ f_{i,j}^{up}(\alpha_k,\beta) \ \oplus \ f_{i,j}^{down}(\alpha_k,\beta)
\nonumber
\\
&
\hspace*{-4mm}
&
\oplus \ f_{i,j}^{left}(\alpha_k,\beta) \ \oplus \ f_{i,j}^{right}(\alpha_k,\beta)\ \big) \times w_{i,j,k}^1 \ \Big] \ ,
\end{eqnarray}
where
$\times$ denotes an element-wise multiplication, $\oplus$ denotes the concatenation operator, and $\oplus_{k=1}^{n}$ concatenates all the feature maps for different attenuation factors, after the feature maps are multiplied with the attention weights ($w_{i,j,k}^1$) by broadcasting the $W_{k}^1$ in a channel-wise manner.
With the attention weights learned to select and integrate the context features aggregated with different attenuation factors (see again Fig.~\ref{fig:motivation}), our network can adaptively control the feature integration and allow the context features to be implicitly dispersed over varying spatial ranges.


\textbf{Completing the SAC module.}
After concatenating the features, we complete the first round of recurrent translations in our SAC module and further apply a $1\times1$ convolution to reduce the feature channels.
Then, we repeat the same process in the second round of recurrent translation using another set of attention weights $\{W^{2}_{1}, W^{2}_{2}, ..., W^{2}_{n}\}$, which are also learnt through the attention mechanism; see again Fig.~\ref{fig:context}.
After two rounds of recurrent translations, each pixel can obtain context features from the global domain adaptively aggregated with different attenuations; see Fig.~\ref{fig:information_flow}(c).
In the end, we further perform a $1\times1$ convolution followed by the group normalization and ReLU non-linear operation on the integrated features to produce the SAC module output, i.e., the spatial attenuation context.
Since two rounds of recurrent translations are able to obtain the global information, we set the number of rounds as two during the experiments.

\begin{figure*}[t]
	\centering
	
	\vspace*{0.5mm}
	\begin{subfigure}{0.115\textwidth}
		\includegraphics[width=\textwidth]{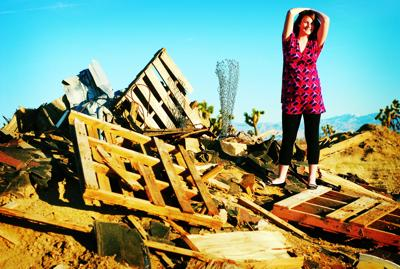}
	\end{subfigure}
	\begin{subfigure}{0.115\textwidth}
		\includegraphics[width=\textwidth]{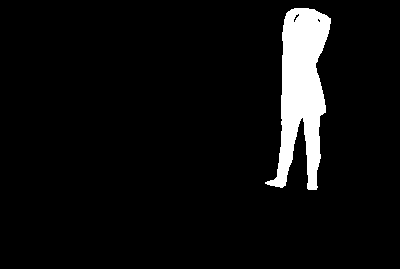}
	\end{subfigure}
	\begin{subfigure}{0.115\textwidth}
		\includegraphics[width=\textwidth]{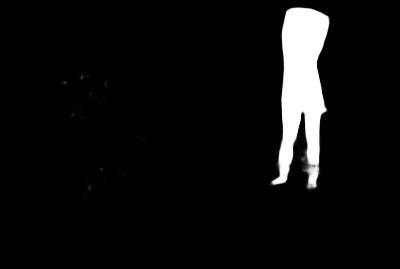}
	\end{subfigure}
	\begin{subfigure}{0.115\textwidth}
		\includegraphics[width=\textwidth]{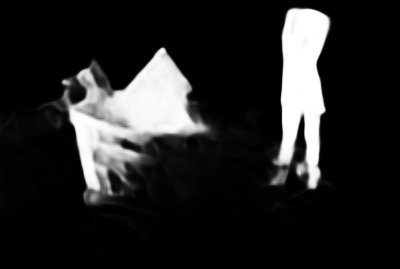}
	\end{subfigure}
	\begin{subfigure}{0.115\textwidth}
		\includegraphics[width=\textwidth]{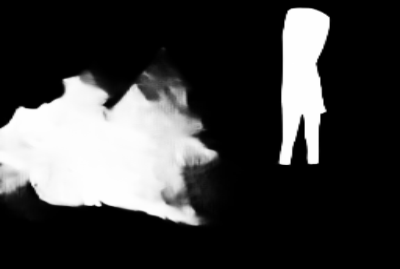}
	\end{subfigure}
	\begin{subfigure}{0.115\textwidth}
		\includegraphics[width=\textwidth]{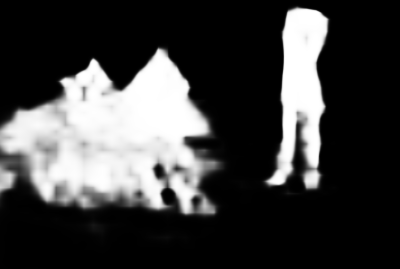}
	\end{subfigure}
	\begin{subfigure}{0.115\textwidth}
		\includegraphics[width=\textwidth]{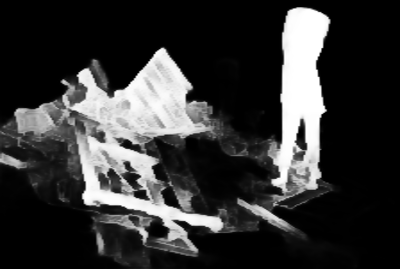}
	\end{subfigure}
	\begin{subfigure}{0.115\textwidth}
		\includegraphics[width=\textwidth]{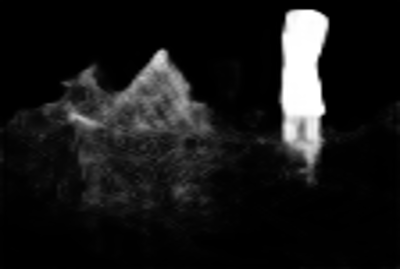}
	\end{subfigure}
	\ \\

		    \vspace*{0.5mm}
	\begin{subfigure}{0.115\textwidth}
		\includegraphics[width=\textwidth]{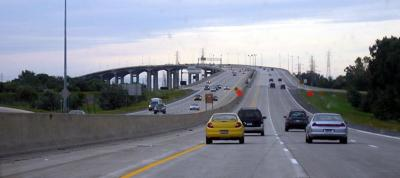}
	\end{subfigure}
	\begin{subfigure}{0.115\textwidth}
		\includegraphics[width=\textwidth]{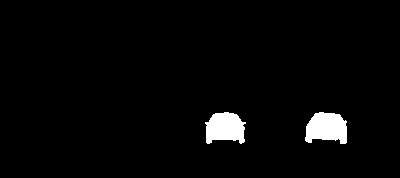}
	\end{subfigure}
	\begin{subfigure}{0.115\textwidth}
		\includegraphics[width=\textwidth]{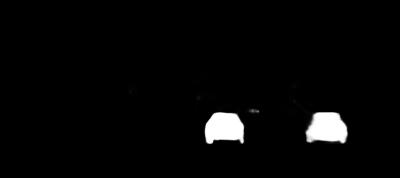}
	\end{subfigure}
	\begin{subfigure}{0.115\textwidth}
		\includegraphics[width=\textwidth]{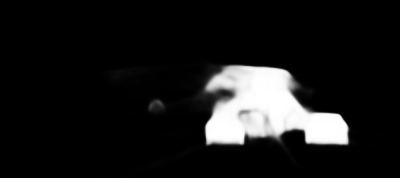}
	\end{subfigure}
	\begin{subfigure}{0.115\textwidth}
		\includegraphics[width=\textwidth]{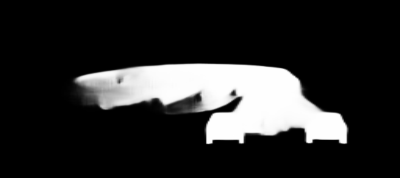}
	\end{subfigure}
	\begin{subfigure}{0.115\textwidth}
		\includegraphics[width=\textwidth]{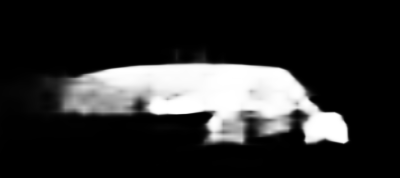}
	\end{subfigure}
	\begin{subfigure}{0.115\textwidth}
		\includegraphics[width=\textwidth]{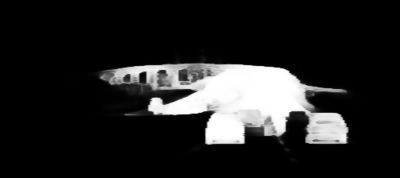}
	\end{subfigure}
	\begin{subfigure}{0.115\textwidth}
		\includegraphics[width=\textwidth]{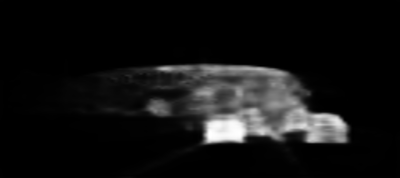}
	\end{subfigure}
	\ \\

			\vspace*{0.5mm}
	\begin{subfigure}{0.115\textwidth}
		\includegraphics[width=\textwidth]{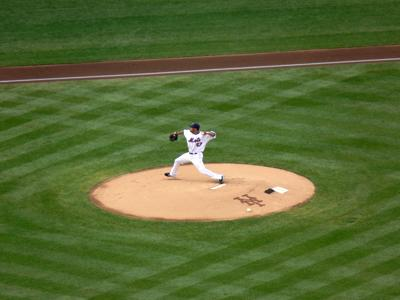}
	\end{subfigure}
	\begin{subfigure}{0.115\textwidth}
		\includegraphics[width=\textwidth]{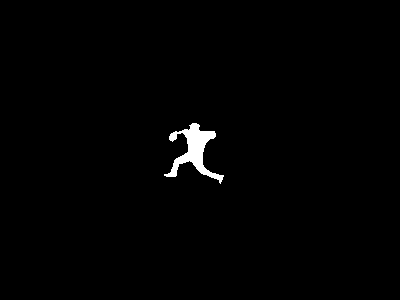}
	\end{subfigure}
	\begin{subfigure}{0.115\textwidth}
		\includegraphics[width=\textwidth]{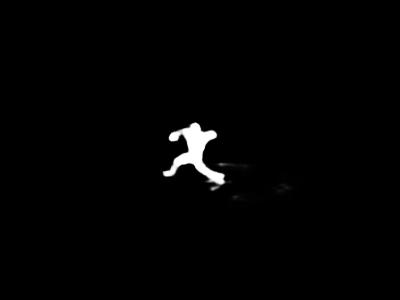}
	\end{subfigure}
	\begin{subfigure}{0.115\textwidth}
		\includegraphics[width=\textwidth]{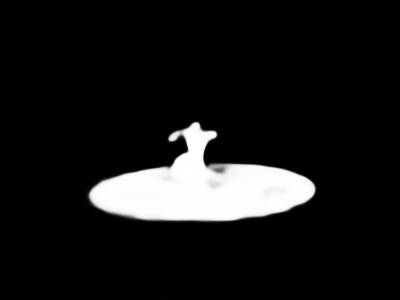}
	\end{subfigure}
	\begin{subfigure}{0.115\textwidth}
		\includegraphics[width=\textwidth]{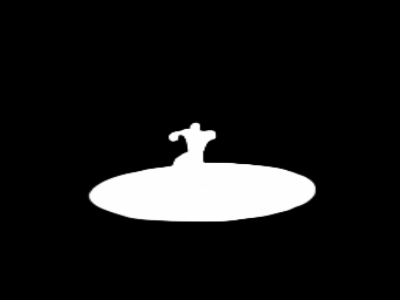}
	\end{subfigure}
	\begin{subfigure}{0.115\textwidth}
		\includegraphics[width=\textwidth]{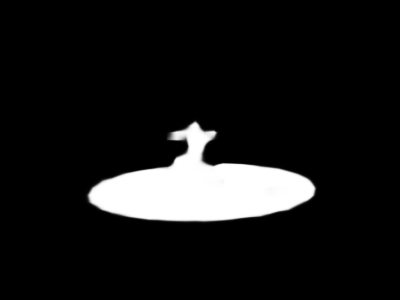}
	\end{subfigure}
	\begin{subfigure}{0.115\textwidth}
		\includegraphics[width=\textwidth]{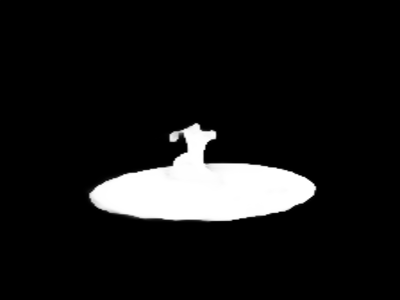}
	\end{subfigure}
	\begin{subfigure}{0.115\textwidth}
		\includegraphics[width=\textwidth]{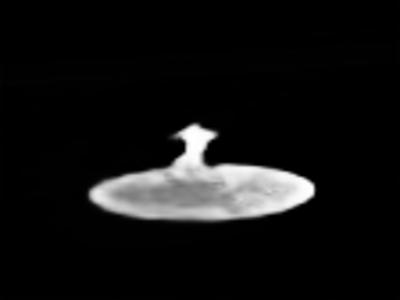}
	\end{subfigure}
	\ \\

	    \vspace*{0.5mm}
	\begin{subfigure}{0.115\textwidth}
		\includegraphics[width=\textwidth]{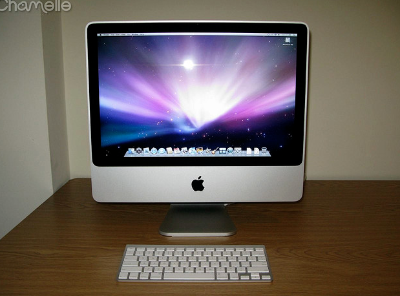}
	\end{subfigure}
	\begin{subfigure}{0.115\textwidth}
		\includegraphics[width=\textwidth]{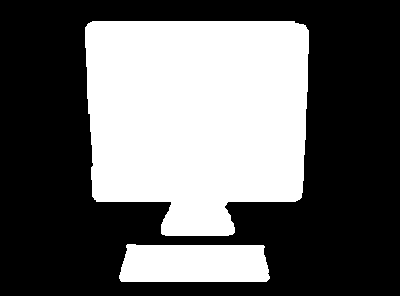}
	\end{subfigure}
	\begin{subfigure}{0.115\textwidth}
		\includegraphics[width=\textwidth]{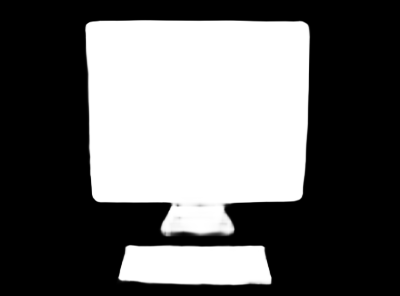}
	\end{subfigure}
	\begin{subfigure}{0.115\textwidth}
		\includegraphics[width=\textwidth]{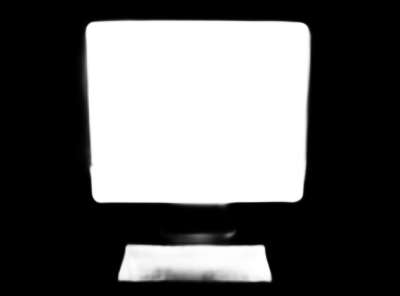}
	\end{subfigure}
	\begin{subfigure}{0.115\textwidth}
		\includegraphics[width=\textwidth]{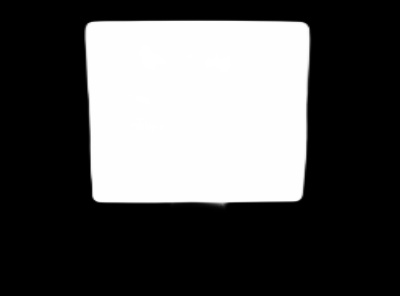}
	\end{subfigure}
	\begin{subfigure}{0.115\textwidth}
		\includegraphics[width=\textwidth]{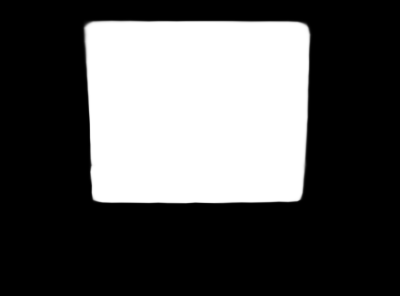}
	\end{subfigure}
	\begin{subfigure}{0.115\textwidth}
		\includegraphics[width=\textwidth]{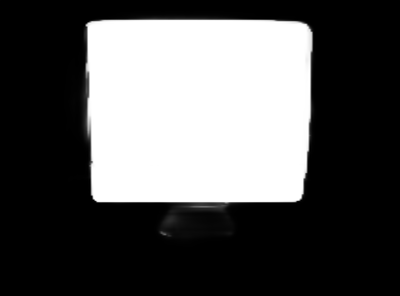}
	\end{subfigure}
	\begin{subfigure}{0.115\textwidth}
		\includegraphics[width=\textwidth]{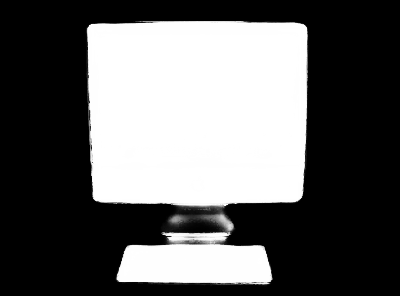}
	\end{subfigure}
	\ \\

	\vspace*{0.5mm}
	\begin{subfigure}{0.115\textwidth}
		\includegraphics[width=\textwidth]{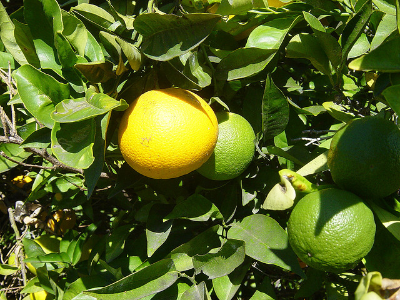}
	\end{subfigure}
	\begin{subfigure}{0.115\textwidth}
		\includegraphics[width=\textwidth]{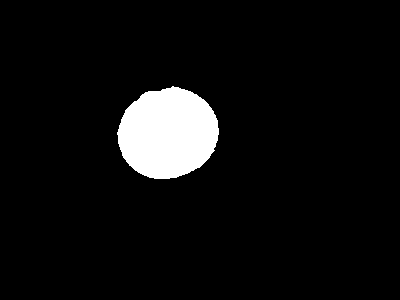}
	\end{subfigure}
	\begin{subfigure}{0.115\textwidth}
		\includegraphics[width=\textwidth]{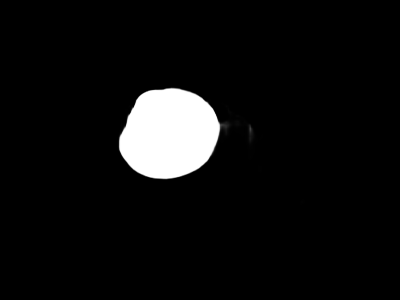}
	\end{subfigure}
	\begin{subfigure}{0.115\textwidth}
		\includegraphics[width=\textwidth]{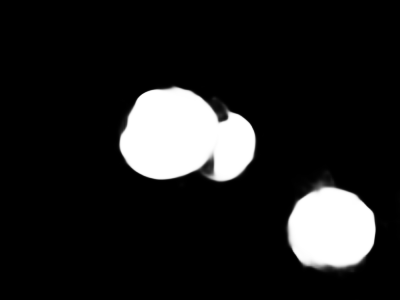}
	\end{subfigure}
	\begin{subfigure}{0.115\textwidth}
		\includegraphics[width=\textwidth]{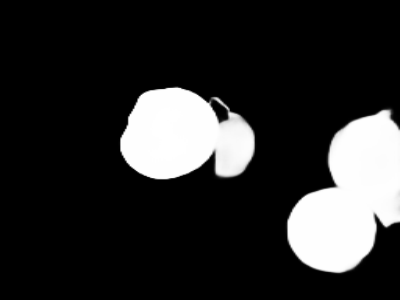}
	\end{subfigure}
	\begin{subfigure}{0.115\textwidth}
		\includegraphics[width=\textwidth]{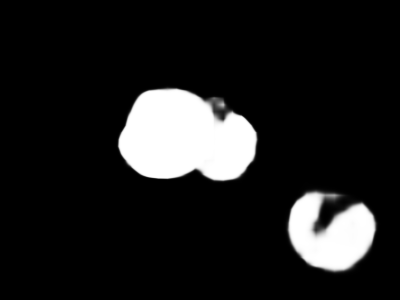}
	\end{subfigure}
	\begin{subfigure}{0.115\textwidth}
		\includegraphics[width=\textwidth]{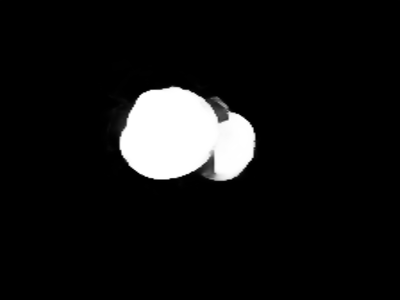}
	\end{subfigure}
	\begin{subfigure}{0.115\textwidth}
		\includegraphics[width=\textwidth]{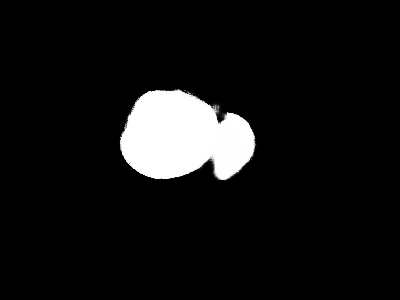}
	\end{subfigure}
	\ \\

	\vspace*{0.5mm}
	\begin{subfigure}{0.115\textwidth}
		\includegraphics[width=\textwidth]{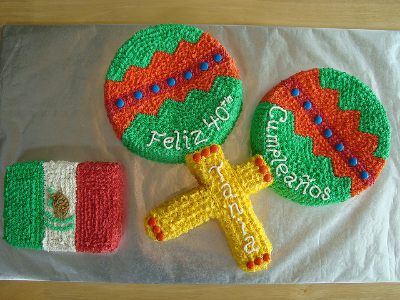}
	\end{subfigure}
	\begin{subfigure}{0.115\textwidth}
		\includegraphics[width=\textwidth]{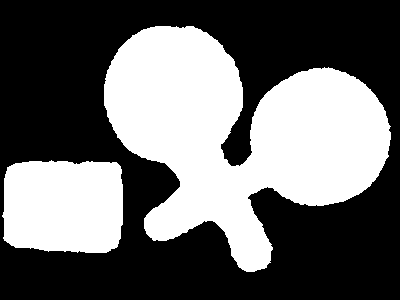}
	\end{subfigure}
	\begin{subfigure}{0.115\textwidth}
		\includegraphics[width=\textwidth]{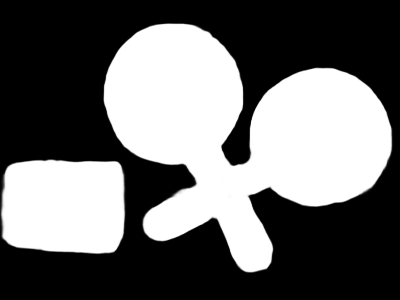}
	\end{subfigure}
	\begin{subfigure}{0.115\textwidth}
		\includegraphics[width=\textwidth]{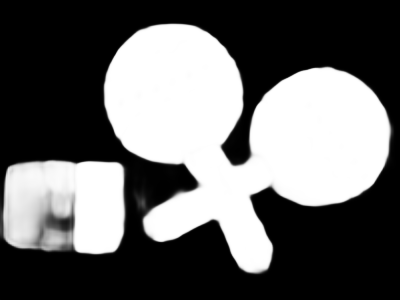}
	\end{subfigure}
	\begin{subfigure}{0.115\textwidth}
		\includegraphics[width=\textwidth]{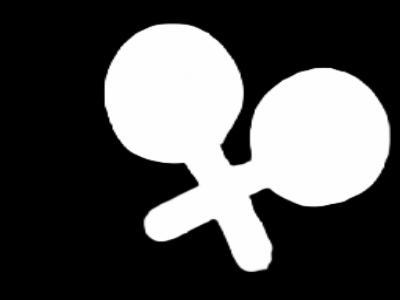}
	\end{subfigure}
	\begin{subfigure}{0.115\textwidth}
		\includegraphics[width=\textwidth]{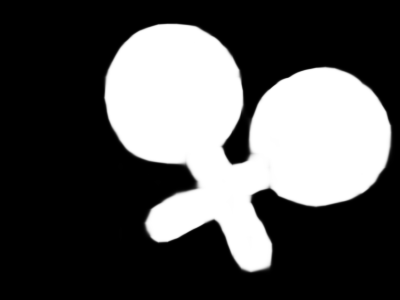}
	\end{subfigure}
	\begin{subfigure}{0.115\textwidth}
		\includegraphics[width=\textwidth]{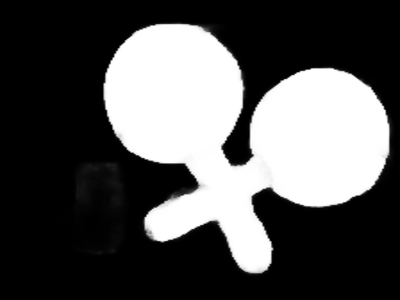}
	\end{subfigure}
	\begin{subfigure}{0.115\textwidth}
		\includegraphics[width=\textwidth]{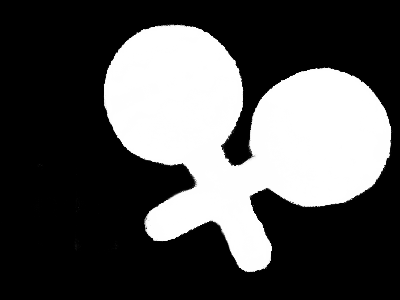}
	\end{subfigure}
	\ \\

	\vspace*{0.5mm}
\begin{subfigure}{0.115\textwidth}
	\includegraphics[width=\textwidth]{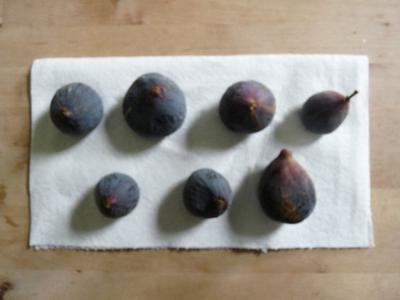}
\end{subfigure}
\begin{subfigure}{0.115\textwidth}
	\includegraphics[width=\textwidth]{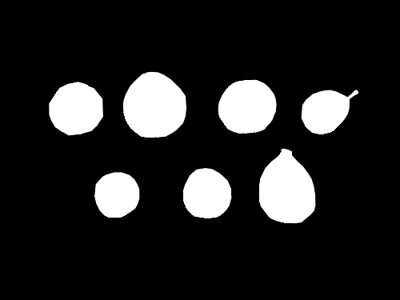}
\end{subfigure}
\begin{subfigure}{0.115\textwidth}
	\includegraphics[width=\textwidth]{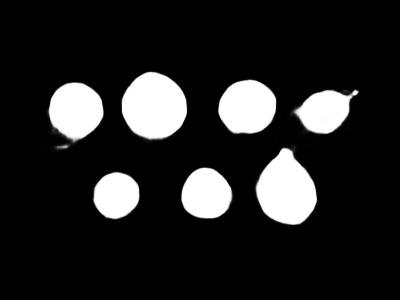}
\end{subfigure}
\begin{subfigure}{0.115\textwidth}
	\includegraphics[width=\textwidth]{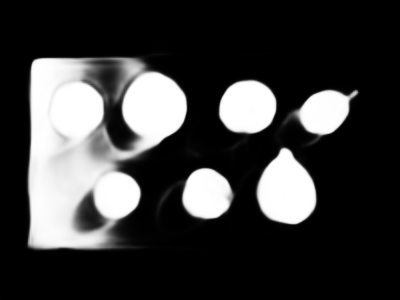}
\end{subfigure}
\begin{subfigure}{0.115\textwidth}
	\includegraphics[width=\textwidth]{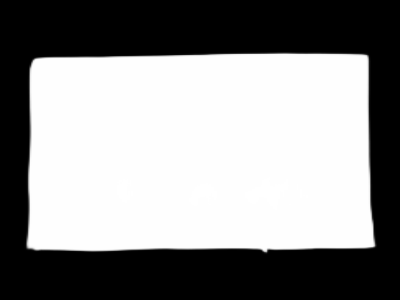}
\end{subfigure}
\begin{subfigure}{0.115\textwidth}
	\includegraphics[width=\textwidth]{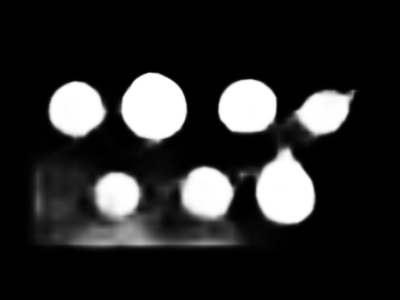}
\end{subfigure}
\begin{subfigure}{0.115\textwidth}
	\includegraphics[width=\textwidth]{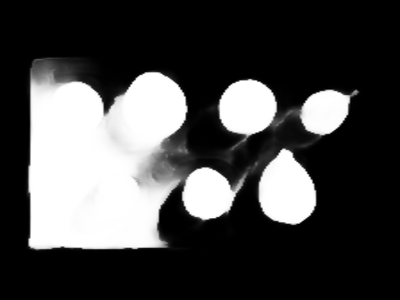}
\end{subfigure}
\begin{subfigure}{0.115\textwidth}
	\includegraphics[width=\textwidth]{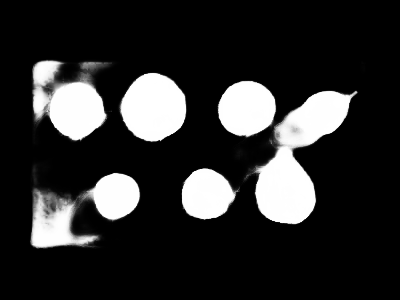}
\end{subfigure}
\ \\

	\vspace*{0.5mm}
\begin{subfigure}{0.115\textwidth}
	\includegraphics[width=\textwidth]{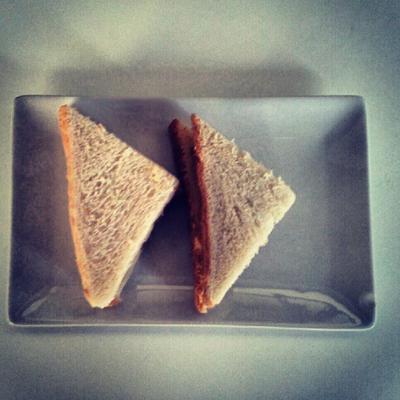}
\end{subfigure}
\begin{subfigure}{0.115\textwidth}
	\includegraphics[width=\textwidth]{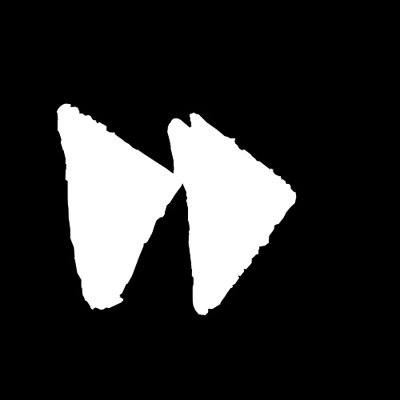}
\end{subfigure}
\begin{subfigure}{0.115\textwidth}
	\includegraphics[width=\textwidth]{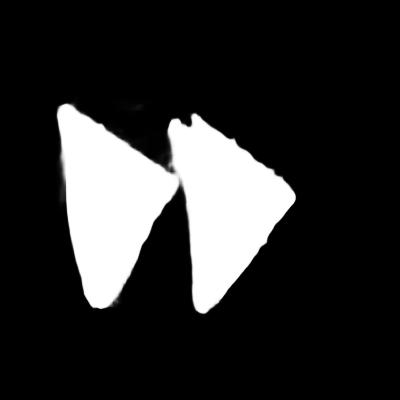}
\end{subfigure}
\begin{subfigure}{0.115\textwidth}
	\includegraphics[width=\textwidth]{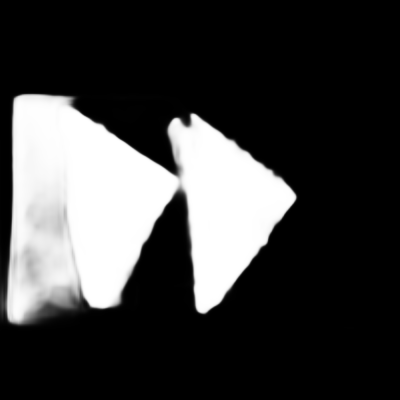}
\end{subfigure}
\begin{subfigure}{0.115\textwidth}
	\includegraphics[width=\textwidth]{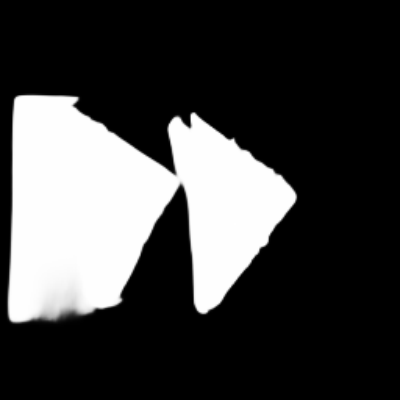}
\end{subfigure}
\begin{subfigure}{0.115\textwidth}
	\includegraphics[width=\textwidth]{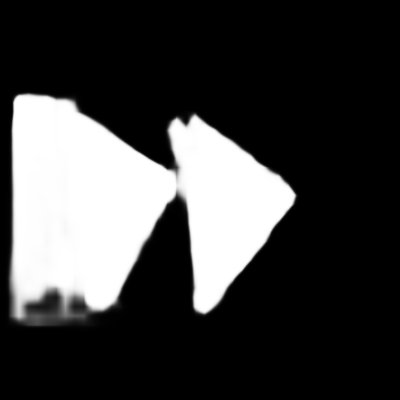}
\end{subfigure}
\begin{subfigure}{0.115\textwidth}
	\includegraphics[width=\textwidth]{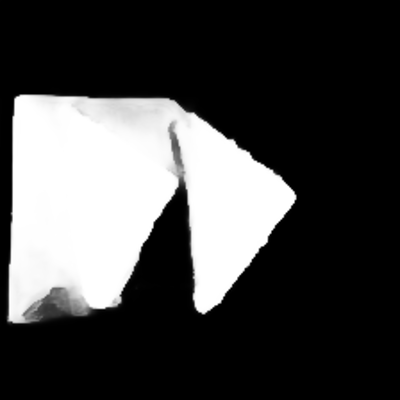}
\end{subfigure}
\begin{subfigure}{0.115\textwidth}
	\includegraphics[width=\textwidth]{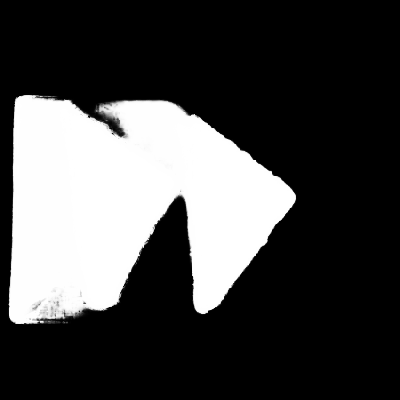}
\end{subfigure}
\ \\

     \vspace*{0.5mm}
     \begin{subfigure}{0.115\textwidth}
     	\includegraphics[width=\textwidth]{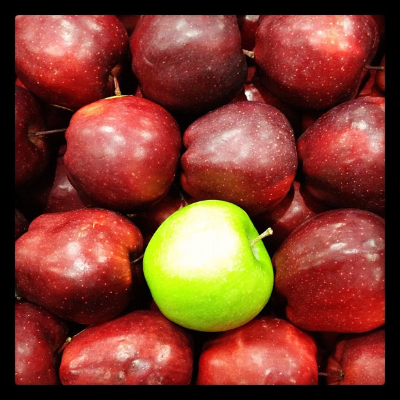}
     	\vspace{-5.5mm} \caption{\footnotesize{inputs}}
     \end{subfigure}
     \begin{subfigure}{0.115\textwidth}
     	\includegraphics[width=\textwidth]{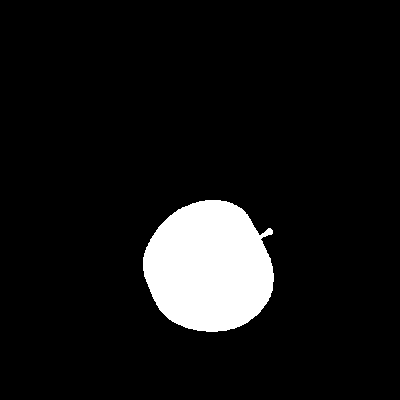}
     	\vspace{-5.5mm} \caption{\footnotesize{ground truth}}
     \end{subfigure}
     \begin{subfigure}{0.115\textwidth}
     	\includegraphics[width=\textwidth]{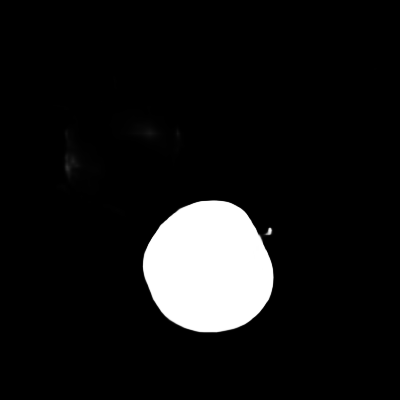}
     	\vspace{-5.5mm} \caption{\footnotesize{ours}}
     \end{subfigure}
     \begin{subfigure}{0.115\textwidth}
     	\includegraphics[width=\textwidth]{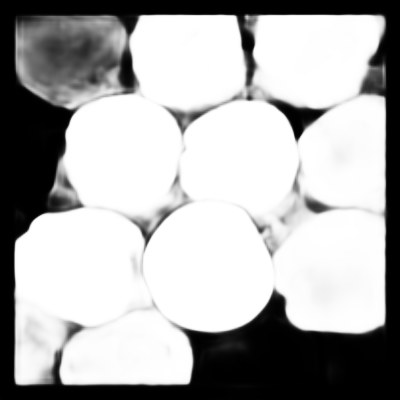}
     	\vspace{-5.5mm} \caption{\scriptsize{PoolNet-R~}\footnotesize{\cite{liu2019simple}}}
     \end{subfigure}
     \begin{subfigure}{0.115\textwidth}
     	\includegraphics[width=\textwidth]{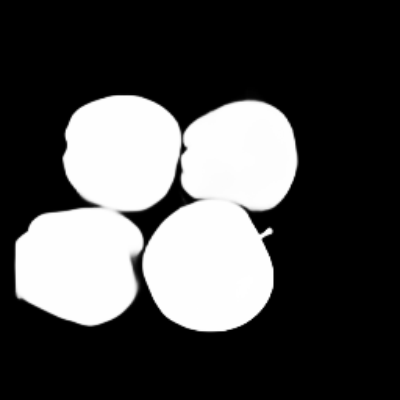}
     	\vspace{-5.5mm} \caption{\footnotesize{BASNet~\cite{qin2019basnet}}}
     \end{subfigure}
     \begin{subfigure}{0.115\textwidth}
     	\includegraphics[width=\textwidth]{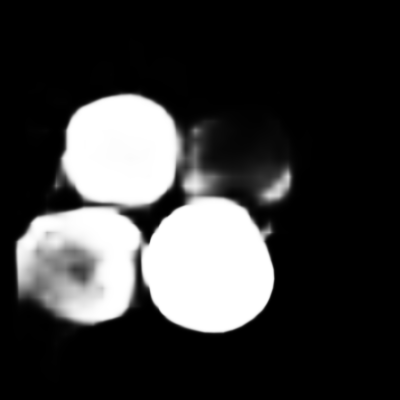}
     	\vspace{-5.5mm} \caption{\footnotesize{CPD-R~\cite{wu2019cascaded}}}
     \end{subfigure}
     \begin{subfigure}{0.115\textwidth}
     	\includegraphics[width=\textwidth]{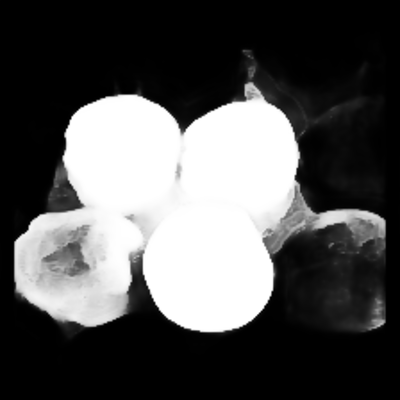}
     	\vspace{-5.5mm} \caption{\footnotesize{AFNet~\cite{feng2019attentive}}}
     \end{subfigure}
     \begin{subfigure}{0.115\textwidth}
     	\includegraphics[width=\textwidth]{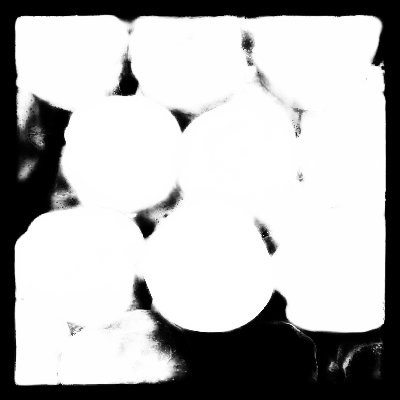}
     		\vspace{-5.5mm} \caption{\scriptsize{PiCA-R~}\footnotesize{\cite{liu2018picanet}}}
     \end{subfigure}
     \ \\

	\caption{Visual comparison of saliency maps (c)-(h) produced by different methods.
Apparently, our method produces more accurate saliency maps. Results are shown before using CRF.}
	\label{fig:comparison_real_photos_part1}
	
\end{figure*}



\subsection{Training and Testing Strategies}
\label{sec:training_testing_strategies}

We built our SAC-Net on ResNet-101~\cite{he2016deep} and used the feature pyramid network (FPN)~\cite{lin2017feature} (green blocks in Fig.~\ref{fig:arc}) to enhance the feature's expressiveness.
Like~\cite{lin2017feature}, we set the channel number of each FPN or SAC layer as $256$ and did not use the feature maps at the first layer in both the FPN or SAC module due to the large memory footprint.
This framework was implemented based on Caffe~\cite{jia2014caffe}.

\textbf{Objective function.}
We used the cross-entropy loss to train the network. Since we have multiple predictions over different layers (from deep to shallow) in our SAC-Net (see Fig.~\ref{fig:arc}), the total loss $L$ is defined as the summation of the cross-entropy loss over all the predicted saliency maps:
\begin{equation}
\label{equ:loss}
L = - \sum_{l} \sum_{i,j} g_{i,j} \log(p_{i,j}^l) - (1 - g_{i,j}) \log(1 - p_{i,j}^l)\ ,
\end{equation}
where
$l$ is the layer index in network,
$g_{i,j}$ is the ground truth value at pixel $(i,j)$ (i.e., one for salient regions, and zero, otherwise), and
$p_{i,j}^l \in [0,1]$ is the predicted saliency value at pixel $(i,j)$ on the result in the network's $l$-th layer.


\textbf{Training parameters.}
We initialized the feature extraction part in our network (frontal blue blocks in Fig.~\ref{fig:arc}) using weights of ResNet-101~\cite{he2016deep} trained on ImageNet~\cite{deng2009imagenet}, and initialized other network parts using random noise.
Moreover, we adopted two different training strategies to optimize the network.
First, we used stochastic gradient descent (SGD) with a momentum value of $0.9$ and a weight decay of $0.0005$, and we set the learning rate as $10^{-8}$, adjusted it to be $10^{-9}$ after $13,000$ training iterations, and stopped the training after $20,000$ iterations.
Second, following~\cite{liu2019simple}, we used Adam~\cite{kingma2015adam} with the first momentum value of $0.9$,
second momentum value of $0.999$, and weight decay of $5\times10^{-4}$. We set the learning rate as $10^{-5}$ and stopped the training after $50,000$ iterations.
The first training strategy is fast while the second strategy achieves better results; see Section~\ref{sec:exp:abl}.
Also, we horizontally flipped the input images for data argumentation in both training strategies.
Lastly, we trained the network on a single NVidia Titan Xp GPU with a mini-batch size of one and updated the weights in every ten training iterations.

\textbf{Inference.}
We took the highest-resolution prediction as the overall result and refined the salient object boundary using fully-connected conditional random field (CRF)~\cite{krahenbuhl2011efficient}.

\begin{table*}  [htbp]
	\begin{center}
		\setlength\tabcolsep{3pt}
		\caption{Comparing our method (SAC-Net) with the state-of-the-art methods using ResNet-101 as the backbone network. Results are reported before using CRF.}
		\label{table:resnet101_state-of-the-art}
		\resizebox{1.0\textwidth}{!}{%
			\begin{tabular}{|c|c|ccc|ccc|ccc|ccc|ccc|ccc|}
				\hline
				Dataset &	-&
				\multicolumn{3}{c|}{ECSSD~\cite{yan2013hierarchical}} &
				\multicolumn{3}{c|}{PASCAL-S~\cite{li2014secrets}} &
				\multicolumn{3}{c|}{SOD~\cite{movahedi2010design}} &
				\multicolumn{3}{c|}{HKU-IS~\cite{li2015visual}} &
				\multicolumn{3}{c|}{DUT-OMRON~\cite{yang2013saliency}} &
				\multicolumn{3}{c|}{DUTS-test~\cite{wang2017learning}} 
				
				\\
				
				\cline{1-20}
				 Metric & Year & $F_\beta$ & $S_m$ & MAE & $F_\beta$ & $S_m$ & MAE & $F_\beta$ & $S_m$ & MAE & $F_\beta$ & $S_m$ & MAE & $F_\beta$ & $S_m$ & MAE & $F_\beta$ & $S_m$ & MAE \\
				\hline
			
				\textbf{SAC-Net (ours)} &- & \textbf{0.951} & \textbf{0.931}& \textbf{0.031} 
				& \textbf{0.879}& \textbf{0.806}& \textbf{0.070} 
				& \textbf{0.882} & \textbf{0.809}& \textbf{0.093} 
				& \textbf{0.942} & \textbf{0.925}& \textbf{0.026}  
				& \textbf{0.830} & \textbf{0.849}& \textbf{0.052} 
				& \textbf{0.895} & \textbf{0.883}& \textbf{0.034} \\

				PoolNet-R+ &2019 & 0.947 &0.924 & 0.032
				& 0.867 & 0.801 & 0.071
				& 0.872 & 0.798 & 0.097
				& 0.937 & 0.919 & \textbf{0.026}
				& 0.813 &0.834 & \textbf{0.052}
				&0.883 &0.873 & 0.035  \\

				BASNet+ & 2019 & 0.919 & 0.894 & 0.049 
				& 0.825 & 0.761 &0.101 
				& 0.825 & 0.754 &0.126
				& 0.912 & 0.893 & 0.040
				& 0.795 & 0.819 & 0.064
				& 0.822 & 0.821 & 0.061 \\

				DSS+~\cite{hou2019deeply} & 2019 & 0.906 & 0.862 & 0.074 & 0.819 & 0.721 & 0.115 & 0.831 & 0.735 & 0.144 & 0.904 & 0.869 & 0.054 & 0.783 & 0.799 & 0.070 & 0.819 & 0.809 & 0.067 \\

				PiCA-R+ &2018 & 0.940 & 0.914 & 0.037
				&0.863&0.791& 0.076
				&0.864&0.768&0.101
				&0.931&0.905&0.031
				&0.816 & 0.828 & 0.068
				&0.868 & 0.844 & 0.043 
				\\

				\hline
					
		\end{tabular} }
	\end{center}
\end{table*}


\section{Experimental Results}
\label{sec:experiment}


\subsection{Datasets and Evaluation Metrics}
We used six widely-used saliency benchmark datasets in our experiments:
(i) ECSSD~\cite{yan2013hierarchical} has $1,000$ natural images with many semantically meaningful but complex structures;
(ii) PASCAL-S~\cite{li2014secrets} has $850$ images generated from the PASCAL VOC2010 segmentation dataset~\cite{everingham2010pascal}, where each image has several salient objects;
(iii) SOD~\cite{movahedi2010design} has $300$ images selected from the BSDS dataset~\cite{martin2001database}, where the salient objects are typically of low contrast or closely contact with the image boundary;
(iv) HKU-IS~\cite{li2015visual} has $4,447$ images, where most images have multiple salient objects;
(v) DUT-OMRON~\cite{yang2013saliency} has $5,168$ high-quality images, each with one or more salient objects; and
(vi) DUTS~\cite{wang2017learning} has a training set of $10,553$ images and a testing set (denoted as DUTS-test) of $5,019$ images, where the images contain various number of salient objects with large variance in scale.
Among the datasets, HKU-IS, DUT-OMRON, and DUTS provide a large number of test images captured under different situations, enabling more comprehensive comparisons among different methods.
Moreover, we follow the recent works on salient object detection~\cite{liu2018picanet,wang2018detect,zhang2018bi,zhang2018progressive} to train our network model using the training set of DUTS~\cite{wang2017learning}.

Next, we used three common metrics for quantitative evaluation: F-measure ($F_\beta$), structure measure ($S_m$) and mean absolute error (MAE).
%
F-measure is a balanced average precision and recall computed from the predicted maps and the ground truth images:
\begin{equation}  \label{eq_F_measure}
F_{\beta} = \frac{(1+\beta^2)\times Precision \times Recall}{\beta^2 \times Precision + Recall},
\end{equation}
where $\beta^2$ is set as $0.3$ to improve the importance of the precision, as suggested in~\cite{achanta2009frequency,hou2017deeply}.
S-measure~\cite{fan2017structure} computes the object-aware and region-aware structural similarity between the predicted map $S$ and ground truth image $G$:
\begin{equation}  \label{eq_S_measure}
S_m= \alpha \times S_o (S, G) + (1-\alpha) \times S_r(S, G) \ ,
\end{equation}
where $S_o$ and $S_r$ denote the object-aware and region-aware structural similarity, respectively;
$\alpha$ is a parameter, which balances the importance of structural similarities, and we followed~\cite{fan2017structure} and set it as $0.5$.
Overall, a large $F_\beta$ or $S_m$ indicates a better result.
MAE~\cite{perazzi2012saliency} is the average pixel-wise absolute difference between the predicted map $S$ and the ground truth image $G$:
\begin{equation}  \label{eq_F_measure}
MAE = \frac{1}{W_{S} \times H_{S}} \sum_{x=1}^{W_{S}}\sum_{y=1}^{H_{S}} \| S(x, y) - G(x, y) \|,
\end{equation}
where $W_{S}$ and $H_{S}$ are the width and height of $S$ or $G$, respectively.
Unlike the $F_\beta$ and $S_m$, a small $MAE$ indicates a better result.
Finally, we used the implementation of~\cite{fan2017structure,hou2017deeply} to compute $F_\beta$, $S_m$ and MAE for all results.


%



\subsection{Comparison with the State-of-the-arts}
\label{subsec:compare_state_of_the_art}

We compared our method with $29$ state-of-the-art methods; see the first column in Table~\ref{table:state-of-the-art}.
Among the methods, to detect salient objects, BSCA~\cite{qin2015saliency} and DRFI~\cite{jiang2013salient} use hand-crafted features, while others employ deep neural networks to learn features.
For a fair comparison, we obtained their results either by using the saliency maps provided by the authors or by producing the results using their implementations with the released training models.

\textbf{Quantitative comparison.}
Table~\ref{table:state-of-the-art} summaries the quantitative results compared with the $29$ state-of-the-art methods in terms of $F_\beta$, $S_m$ and MAE on detecting salient objects in the six benchmark datasets.
Our SAC-Net performs favorably against all the others for almost all the cases, regardless of whether CRF is used as a post-processing step.
Especially, our method without CRF (SAC-Net) already achieves the best performance compared with all the other methods with CRF for most datasets.
This result demonstrates the strong capability of our method to deal with challenging inputs; see also the visual comparison results presented in Fig.~\ref{fig:comparison_real_photos_part1}.

Recent deep learning methods use different kinds of backbone networks for feature extraction. For a fair comparison, we retrained these methods (PoolNet~\cite{liu2019simple}, BASNet~\cite{qin2019basnet}, and PiCA~\cite{liu2018picanet}) by using the same backbone network (ResNet-101) as our SAC-Net. We reported the results of DSS~\cite{hou2019deeply} using ResNet-101 by downloading the trained model from the authors' website. These models are denoted as ``XX+''. Table~\ref{table:resnet101_state-of-the-art} shows the comparison results, where our method still outperforms the very recent salient object detection methods on all the benchmark datasets.
We also re-train our method by taking ResNet-50 as the backbone network, and report the results ``SAC-Net (Res50)'' in the last row of Table~\ref{table:state-of-the-art}, where our method still achieves the best performance on most of the benchmark datasets.


\begin{table*}  [htbp]
	\begin{center}
		\setlength\tabcolsep{3pt}
		\caption{Component analysis.
Note that ``SC'' denotes ``spatial context,'' ``TS'' denotes ``training strategy,''  
and ``with LSTM'' denotes the use of long short-term memory to aggregate the spatial context features.}
		\label{table:component_analysis}
		\resizebox{1.0\textwidth}{!}{
			\begin{tabular}{|c|c|c|ccc|ccc|ccc|ccc|ccc|ccc|}
				\hline
				&\multirow{2}{*}{SC} & \multirow{2}{*}{TS} &
				
				\multicolumn{3}{c|}{ECSSD} &	\multicolumn{3}{c|}{PASCAL-S}  & \multicolumn{3}{c|}{SOD} &  \multicolumn{3}{c|}{HKU-IS} & \multicolumn{3}{c|}{DUT-OMRON} & \multicolumn{3}{c|}{DUTS-test}  \\
				
				\cline{4-21}
				&&& $F_\beta$ & $S_m$ & MAE & $F_\beta$ & $S_m$ & MAE & $F_\beta$ & $S_m$ & MAE & $F_\beta$ & $S_m$ & MAE & $F_\beta$ & $S_m$ & MAE & $F_\beta$ & $S_m$ & MAE \\
				\hline
				FPN~\cite{lin2017feature}&$\times$ & SGD & 0.926 & 0.904 & 0.056 
				& 0.859& 0.780& 0.085 
				& 0.846 & 0.772 &0.124 
				&0.913 & 0.898 &0.046 
				& 0.805 & 0.825& 0.065 
				& 0.858 & 0.852& 0.052 \\
				\hline
				
				\multirow{2}{*}{SAC-Net} &$\surd$ & SGD & {0.949} & {0.928}& {0.036} 
				& {0.878}& {0.805}& {0.072} 
				& {0.874} & {0.806}& {0.099} 
				& {0.938} & {0.923}& {0.030}  
				& {0.828} & \textbf{0.849}& {0.055} 
				& {0.888} & {0.879}& {0.038} \\
				
				\cline{2-21}
				&$\surd$ & Adam & \textbf{0.951} & \textbf{0.931}& \textbf{0.031} 
				& \textbf{0.879}& \textbf{0.806}& \textbf{0.070} 
				& \textbf{0.882} & \textbf{0.809}& \textbf{0.093} 
				& \textbf{0.942} & \textbf{0.925}& \textbf{0.026}  
				& \textbf{0.830} & \textbf{0.849}& \textbf{0.052} 
				& \textbf{0.895} & \textbf{0.883}& \textbf{0.034} \\

				
				\hline
				
				with LSTM &$\surd$ & SGD & 0.941& 0.920 &0.040 
				& 0.872 & 0.794& 0.074 
				& 0.860 & 0.778& 0.111 
				& 0.930 & 0.912& 0.034 
				& 0.825 & 0.836& 0.054  
				& 0.881 & 0.871& 0.040 \\
				
				\hline
				
		\end{tabular} }
	\end{center}
\vspace{-1mm}
\end{table*}



\textbf{Visual comparison.}
Fig.~\ref{fig:comparison_real_photos_part1} presents salient object detection results produced by various methods, including ours.
From the figures, we can see that other methods (d)-(h) tend to include non-salient backgrounds or miss some salient details, while our SAC-Net is able to produce results (c) that are more consistent with the ground truth images (b).
Particularly, for challenging cases, such as
(i) salient objects and non-salient background with similar appearance (see $2^{nd}$ and $4^{th}$ rows),
(ii) small salient objects (see $2^{nd}$ and $3^{rd}$ rows),
(iii) complex background (see $1^{st}$, $2^{nd}$, $4^{th}$, $5^{th}$, and $9^{th}$ rows), and
(iv) multiple objects (see $2^{nd}$, $4^{th}$, and $6^{th}$ to $8^{th}$ rows),
our method can still predict more plausible saliency maps than the others, showing the robustness and quality of SAC-Net.


\subsection{Evaluation on the Network Design} \label{sec:exp:abl}


\textbf{Component analysis.}
We performed an ablation study to evaluate the major components in SAC-Net.
The first row of Table~\ref{table:component_analysis} shows the results from a basic model (FPN~\cite{lin2017feature}) built with only the feature pyramid; see the green blocks in Fig.~\ref{fig:arc}.
By having the SAC modules in the network to adaptively aggregate spatial context, we can see clear improvements on all the benchmark datasets as compared with the FPN results; see the first two rows in the table.

\textbf{Training strategy analysis.} As mentioned in Section~\ref{sec:training_testing_strategies}, we adopted two different training strategies to optimize the network.
The second and third rows in Table~\ref{table:component_analysis} show the comparison results, where using Adam achieves better results than using SGD.
However, ``Adam'' took around $45$ hours to train the model, while ``SGD'' took only around $15$ hours.
Hence, we adopted ``SGD'' to perform the following experiments to evaluate network design.

\begin{table}[!t]
	\begin{center}
		\setlength\tabcolsep{2pt}
		\caption{Parameter analysis of SAC module. ``$n$'' is the number of attenuation factors and $\beta$ is defined in Eq.~\eqref{eq:propagation}; see Sec.~\ref{subsec:SACmodule}. Results are reported before using CRF.}
		\label{table:param_analysis}
		\resizebox{1.0\linewidth}{!}{
			\begin{tabular}{|c|c|ccc|ccc|ccc|}
				\hline
				\multirow{2}{*}{$n$} & \multirow{2}{*}{$\beta$} &
				\multicolumn{3}{c|}{HKU-IS} & \multicolumn{3}{c|}{DUT-OMRON} &
				\multicolumn{3}{c|}{DUTS-test} \\
				
				\cline{3-11}
			     &	& $F_\beta$ & $S_m$ & MAE & $F_\beta$ & $S_m$ & MAE & $F_\beta$ & $S_m$ & MAE \\
				\hline
				1 & learnable & 0.928 & 0.914 & 0.035 & 0.824 & 0.836 & 0.058 & 0.875 & 0.866 & 0.043 \\
				2 & learnable & 0.937 & 0.921 & 0.031 & 0.826 & 0.843 & 0.057 & 0.886 & 0.877 & 0.039 \\
				3 & learnable & \textbf{0.938} & \textbf{0.923} & \textbf{0.030} & 0.828 & \textbf{0.849} & \textbf{0.055} & \textbf{0.888} & \textbf{0.879} & \textbf{0.038} \\
				4 & learnable & 0.937 & 0.922 & \textbf{0.030} & \textbf{0.829} & 0.846 & 0.056 & \textbf{0.888} & 0.878 & \textbf{0.038}  \\
				5 & learnable & 0.937 & 0.921 & 0.031 & 0.825& 0.844 & 0.057&  0.887 & 0.878 & 0.039 \\
				\hline
				3 & fixed ($0.1$) & 0.936 & 0.921 & 0.031 & 0.825& 0.846 &0.056&  0.887 & 0.877 & 0.039 \\

				3 & fixed ($0$) & 0.936 & 0.922 & \textbf{0.030} & 0.824 & 0.844 & 0.058 & 0.883 & 0.875 & 0.040 \\
				3 & fixed ($1$) & 0.935 & 0.920 & 0.032 & 0.826 & 0.846 & 0.057 & 0.884 & 0.875 & 0.041\\
				\hline	
							
		\end{tabular} }
	\end{center}
\vspace{-1mm}
\end{table}

\textbf{Compare with LSTM.}
The long short-term memory~\cite{hochreiter1997long} (LSTM) is an efficient recurrent neural network to process sequence data by using a set of gates.
The method has been extended to process 2D spatial information by some recent works on image classification~\cite{visin2015renet} and saliency detection (s.t., DSCLRCN~\cite{liu2018deep} and PiCA~\cite{liu2018picanet}).
We performed another experiment by adopting the LSTMs in four principal directions with two rounds of recurrent translations to replace our recurrently-attenuating translation model in the SAC module; in detail, we replaced the feature maps with colored arrows in Fig.~\ref{fig:context} by the LSTMs in corresponding directions.

The last row in Table~\ref{table:component_analysis} presents the LSTM results.
Comparing with our results in the second row, we can see that our method performs better for $F_\beta$, $S_m$ and MAE on all the benchmark data.
We think the reason is that due to the limitation of the gate functions in LSTM~\cite{le2015simple}, context features can only propagate over a short distance, thus limiting the dispersal of local context features in the spatial domain.
%
On the other hand, the time complexity of computing LSTMs on 2D feature maps is very high. ``with LSTM'' took around $213$ hours to train the model, while our method took only around $15$ hours, which is more than $14$ times faster.

\textbf{Parameter analysis.}
To build our network, we empirically determine the value of $n$, which affects the number of attenuation factors and the number of feature channels in each aggregated feature map ($\floor*{\frac{256}{n}}$); see Fig.~\ref{fig:context}.
%
In general, a large $n$ allows the network to consider more variety of attenuation factors but each feature map would capture less information in return, since we keep the overall memory consumption to be manageable.
Another parameter in our network is $\beta$, where we automatically learn its value for regulating the magnitude of the negative part in Eq.~\eqref{eq:propagation}.

We evaluated our network on the three largest datasets (HKU-IS, DUT-OMRON, and DUTS-test) using different $n$ and learnable/fixed $\beta$.
The results shown in Table~\ref{table:param_analysis} reveal that when we aggregate the image context using two different attenuation factors ($n$$=$$2$), we achieve better results than using only one single long-range aggregation ($n$$=$$1$).
The results further improve with larger $n$ and roughly stabilizes when $n$ reaches three, so we set $n$$=$$3$.
On the other hand, comparing the results on the 3rd and last three rows (all with $n$$=$$3$) in table, we can see that
automatically learning and adjusting $\beta$ gives better results than using a fixed $\beta$ ($\beta = 0.1 \ \text{or} \ 0$), or linearly aggregating the spatial features ($\beta = 1$).

\begin{table}[!t]
	\begin{center}
		\setlength\tabcolsep{2pt}
		\caption{Architecture analysis of SAC module. Results are reported before using CRF.}
		\label{table:architecture_analysis}
		\resizebox{1.0\linewidth}{!}{
			\begin{tabular}{|c|ccc|ccc|ccc|}
				\hline
				\multirow{2}{*}{Models} &
				\multicolumn{3}{c|}{HKU-IS} & \multicolumn{3}{c|}{DUT-OMRON} &
				\multicolumn{3}{c|}{DUTS-test} \\
				
				\cline{2-10}
			     &	$F_\beta$ & $S_m$ & MAE & $F_\beta$ & $S_m$ & MAE & $F_\beta$ & $S_m$ & MAE \\
				\hline

			    one-round & 0.933 & 0.920 & 0.032 & 0.824 & 0.845 & 0.056 & 0.883 & 0.875 & 0.040 \\
				three-round & 0.937& 0.922 & 0.031 & \textbf{0.828} & 0.848 & 0.055 & 0.886 & \textbf{0.879} & 0.039 \\
				\hline
			    w/o left-right & 0.935 & 0.920 & 0.031 & 0.823 & 0.843 & 0.057 & 0.885 & 0.876 & 0.039 \\
			    w/o up-down & 0.936 & 0.921 & 0.031 & 0.824 & 0.843 & 0.056 & 0.886 & 0.877 & 0.039  \\
			    \hline
			
			    w/o attention & 0.934 & 0.919 & 0.032 & 0.824 & 0.844 & 0.055 & 0.884 & 0.876 & 0.039 \\

			    \hline

				Ours & \textbf{0.938} & \textbf{0.923} & \textbf{0.030} & \textbf{0.828} & \textbf{0.849} & \textbf{0.055} & \textbf{0.888} & \textbf{0.879} & \textbf{0.038} \\

				\hline	
							
		\end{tabular} }
	\end{center}
\vspace{-1mm}
\end{table}


\begin{table*}  [htbp]
	\begin{center}
		\setlength\tabcolsep{3pt}
		\caption{Comparing our method (SAC-Net) with other works on spatial context using ResNet-101 as the backbone network. Results are reported before using CRF.}
		\label{table:non-local}
		\resizebox{1.0\textwidth}{!}{%
	\begin{tabular}{|c|ccc|ccc|ccc|ccc|ccc|ccc|}
		\hline
	    Dataset &
	    \multicolumn{3}{c|}{ECSSD~\cite{yan2013hierarchical}} &
	    \multicolumn{3}{c|}{PASCAL-S~\cite{li2014secrets}} &
	    \multicolumn{3}{c|}{SOD~\cite{movahedi2010design}} &
	    \multicolumn{3}{c|}{HKU-IS~\cite{li2015visual}} &
	    \multicolumn{3}{c|}{DUT-OMRON~\cite{yang2013saliency}} &
	    \multicolumn{3}{c|}{DUTS-test~\cite{wang2017learning}} 
	    
		\\
		
		\cline{1-19}
		Metric & $F_\beta$ & $S_m$ & MAE & $F_\beta$ & $S_m$ & MAE & $F_\beta$ & $S_m$ & MAE & $F_\beta$ & $S_m$ & MAE & $F_\beta$ & $S_m$ & MAE & $F_\beta$ & $S_m$ & MAE \\
		\hline
		
		\textbf{SAC-Net (ours)} & \textbf{0.951} & \textbf{0.931}& \textbf{0.031} 
		& \textbf{0.879}& \textbf{0.806}& \textbf{0.070} 
		& \textbf{0.882} & \textbf{0.809}& \textbf{0.093} 
		& \textbf{0.942} & \textbf{0.925}& \textbf{0.026}  
		& \textbf{0.830} & \textbf{0.849}& \textbf{0.052} 
		& \textbf{0.895} & \textbf{0.883}& \textbf{0.034} \\

	DSC~\cite{Hu_2018_CVPR,hu2019direction} & 0.948 & 0.929 & 0.036
	& 0.877 & 0.801 & 0.072
	& 0.872 & 0.801 & 0.100
	& 0.935 & 0.920 & 0.031
	& \textbf{0.830} & 0.847 & 0.053 
	& 0.886 & 0.878 & 0.038 \\

	DeepLabv3+~\cite{chen2018deeplab} & 0.947 & 0.925 & 0.037
	& 0.878 & 0.797 & 0.071
	& 0.862 & 0.788 & 0.102
	& 0.934 & 0.915 & 0.032
	& 0.824 & 0.836 & 0.053
	& 0.885 & 0.872 & 0.038
	\\

	PSPNet~\cite{zhao2017pyramid} & 0.940 & 0.917 & 0.042
	& 0.877 & 0.795 & 0.071
	& 0.860 & 0.777 & 0.111
	& 0.927 & 0.911 & 0.036
	& 0.819 & 0.829 & 0.056
	& 0.881 & 0.869 & 0.040 \\
	
	PSANet~\cite{zhao2018psanet} & 0.940 & 0.917 & 0.042
	& 0.873 & 0.796 & 0.073
	& 0.858 & 0.778 & 0.112
	& 0.928 & 0.912 & 0.036
	& 0.816 & 0.831 & 0.056
	& 0.879 & 0.869 & 0.041\\
	
	DeepLabv3~\cite{chen2017rethinking} & 0.939 & 0.917 & 0.042
	& 0.873 & 0.793 & 0.073
	& 0.862 & 0.775 & 0.111
	& 0.926 & 0.909 & 0.037 
	& 0.821 & 0.827 & 0.056
	& 0.877 & 0.865 & 0.042 \\
	
	Non-local Network~\cite{wang2018non} & 0.936 & 0.915 & 0.044
	&  0.874 & 0.795 & 0.072
	& 0.858 & 0.776 & 0.112
	& 0.924 & 0.906 & 0.037
	& 0.809 & 0.826 & 0.059 
	&0.873 & 0.865 & 0.043 
	\\
			
	 DeepLab~\cite{chen2015semantic} & 0.934 & 0.914 & 0.045
	 & 0.873 & 0.797 & 0.073
	 & 0.850 & 0.774 & 0.114
	 & 0.923 & 0.907 & 0.038
	 & 0.803 & 0.822 & 0.059
	 & 0.871 & 0.863 & 0.043 \\

		\hline
		
\end{tabular} }
\end{center}
\end{table*}

\textbf{Architecture analysis.}
To evaluate the effectiveness of our network design, we construct several variant models of our network.
As shown in the Table~\ref{table:architecture_analysis}, first, we replace the two-round recurrent translations in our SAC module by one-round and three-round.
The results show that our method with two-round recurrent translations achieves the best performance.
Then, we build two modules, i.e., ``w/o left-right'' and ``w/o up-down'', by removing the recurrent translations in the left and right or up and down directions, which leads to the worse results.
Finally, we remove the attention mechanism in the SAC module to build ``w/o attention''.
Results show that our network design achieves the best performance.
%

\begin{figure} [tp]
	\centering
	\includegraphics[width=0.99\linewidth]{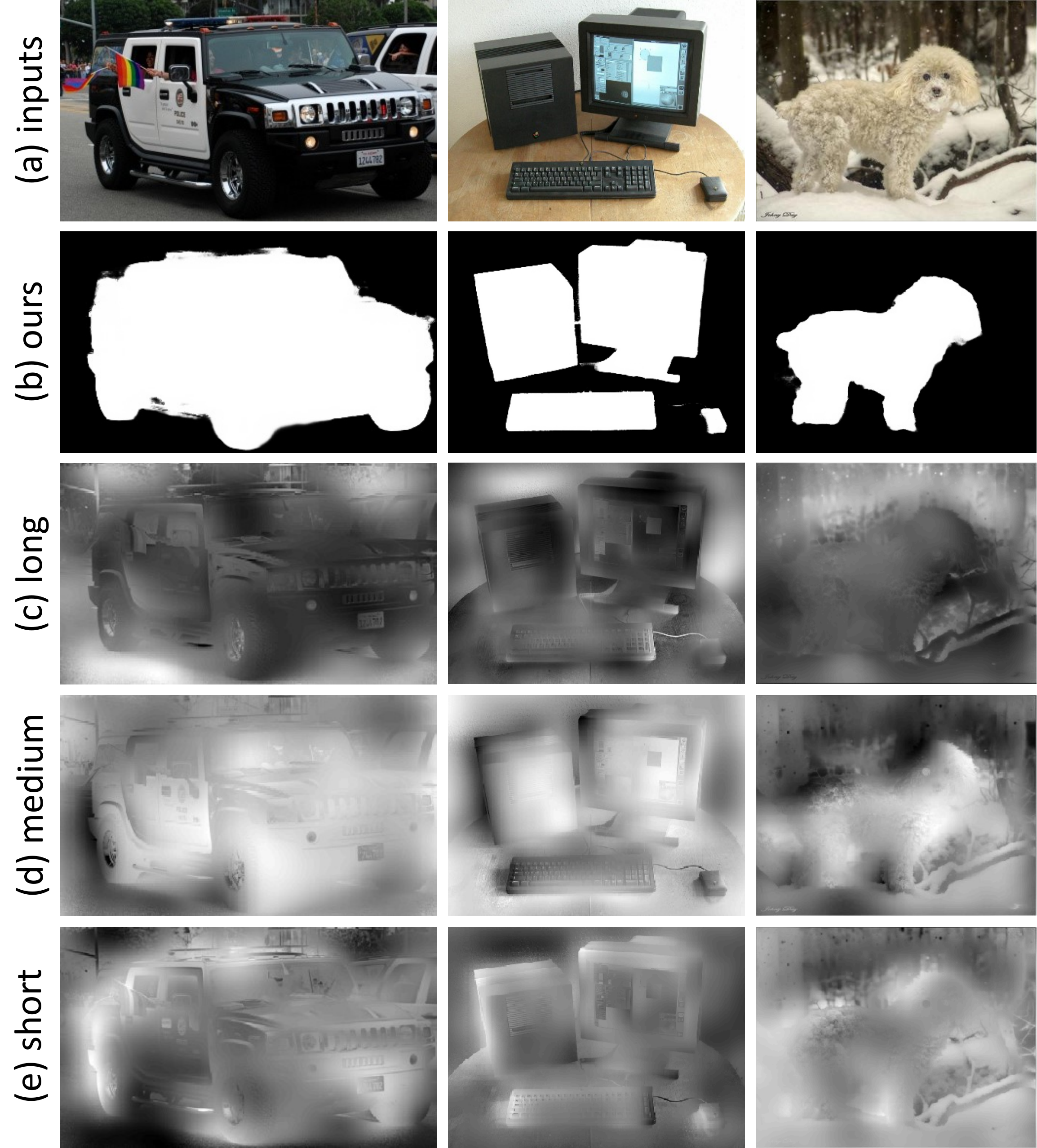}
	\caption{Attention weights learned for different spatial ranges, where the brightness indicates the magnitude of the learned attention weights.}
	\label{fig:attention_weights}
\end{figure}


\textbf{Attention weight visualization.} Figs.~\ref{fig:motivation}~\&~\ref{fig:attention_weights} visualize
the learned attention weights for integrating the spatial context
features.
The long-range context (c) helps to locate the global background regions; the medium-range context (d) helps to identify the image regions of objects;
and short-range context (e) helps to locate the boundary between salient and non-salient regions.
Moreover, our attention mechanism selectively aggregates various spatial context and allows the context features to be implicitly dispersed over arbitrary spatial ranges.

\begin{table}[!t]
	\begin{center}
		\setlength\tabcolsep{2pt}
		\caption{Time performance analysis. ``FPS'' stands for ``frames per second.''}
		\label{table:time_performace}
		\resizebox{0.85\linewidth}{!}{
			\begin{tabular}{|c|c|c|c|c|c|c|c|}
				\hline
				
				Method & Ours & PiCANet~\cite{liu2018picanet} & DGRL~\cite{wang2018detect}  & R$^3$Net~\cite{deng18r} \\
				\hline
				FPS & 11 & 7 & 8 & 4 \\
				
				\hline
				
				Method & SRM~\cite{wang2017stagewise} & Amulet\cite{zhang2017amulet} & NLDF~\cite{luonon2017} & DSS~\cite{hou2017deeply}  \\
				\hline
				FPS & 14 & 16 & 12 & 12  \\
				
				\hline	
				
		\end{tabular} }
	\end{center}

\end{table}

\textbf{Time performance.}
Our network is fast, since it has a fully convolutional architecture and employs an efficient recurrent translation module. We tested our network on a single GPU (TITAN Xp) using input images of size $400\times400$.
It takes around $0.090$ seconds on average to test one image.
If we remove the SAC modules from our network, it still needs $0.087$ seconds to process one image, which proves the efficiency of the proposed SAC module.
Moreover, we compare the time performance of our SAC-Net with other methods for salient object detection.  Table~\ref{table:time_performace} shows the results, where our method has comparable time performance with other methods that have worse detection accuracy than ours.

\begin{table}[!t]
	\begin{center}
		\caption{Comparing with state-of-the-art methods on shadow detection. All the deep-learning methods are trained on the SBU training set and tested on the SBU testing set.}
		\resizebox{0.8\linewidth}{!}{%
		\label{table:shadow}
		\begin{tabular}{|c|c|c|}
			
			\hline
			\multicolumn{2}{|c|}{Method} & BER
			\\
			\hline
		    \multirow{6}{*}{shadow detection} &  BDRAR~\cite{zhu2018bidirectional} & \textbf{3.64}
			\\
			&DSC~\cite{Hu_2018_CVPR,hu2019direction} & 5.59
			\\
			&scGAN~\cite{nguyen2017shadow} & 9.10
			\\
			&stacked-CNN~\cite{vicente2016large} & 11.00
			\\
			&patched-CNN~\cite{hosseinzadeh2018fast} & 11.56
			\\
			&Unary-Pairwise~\cite{guo2011single} & 25.03
			\\
			\hline
			\multirow{7}{*}{saliency detection} & \textbf{SAC-Net (ours)} & \textbf{4.71} \\
			&R$^3$Net~\cite{deng18r} & 5.21 \\
			& PiCANet~\cite{liu2018picanet} & 5.75 \\
			&RADF~\cite{Hu_2018_AAAI} & 6.02 \\
			&SRM~\cite{wang2017stagewise} & 7.25
			\\
			&RAS~\cite{chen2018reverse} & 7.31 \\
			&Amulet~\cite{zhang2017amulet} & 15.13
			\\
			\hline
		\end{tabular} }
	\end{center}
\end{table}

\subsection{Shadow Detection}
Our SAC model has the potential to be applied to other vision tasks. Here, we take
the shadow detection as an example.  We re-train our network as well as other salient object detection methods on the training set of SBU~\cite{vicente2016large}, which is a widely used dataset for shadow detection, and test them on the testing set of SBU. Moreover, we use the common metric BER for the quantitative comparisons among different shadow detectors.
Table~\ref{table:shadow} reports the results, where our SAC-Net achieves the best performance among the methods designed for salient object detection and also outperforms most of the shadow detection methods.

\section{Discussion}
There has been a lot of works on exploiting spatial context in deep CNNs for image analysis.
Dilated convolution~\cite{chen2015semantic,yu2016multi} takes context from larger regions by inserting holes into the convolution kernels, but the context information in use still has a fixed range in a local region.
ASPP~\cite{chen2018deeplab,chen2017rethinking} and PSPNet~\cite{zhao2017pyramid} adopt multiple convolution kernels with different dilated rates or multiple pooling operations with different scales to aggregate spatial context using different region sizes; however, their designed kernel or pooling sizes are fixed, less flexible, and not adaptable to different inputs.
DSC~\cite{Hu_2018_CVPR,hu2019direction} adopts the attention weights to indicate the importance of context features aggregated from different directions, but it only obtains the global context with a fixed influence range over the spatial domain.
The non-local network~\cite{wang2018non} computes correlations between every pixel pair on the feature map to encode the global image semantics, but this method ignores the spatial relationship between pixels in the aggregation; for salient object detection, features of opposite semantics may, however, be important; see Fig.~\ref{fig:motivation}.
PSANet~\cite{zhao2018psanet} adaptively learns attention weights for each pixel to aggregate the information from different positions; however, it is unable to capture the context on lower-level feature maps in high resolutions due to the huge time and memory overhead.
Compared to these methods, our SAC-Net explores and adaptively aggregates context features implicitly with variable influence ranges; it is flexible, fast, and computationally friendly for efficient salient object detection.

We performed an experiment by training these methods on the DUTS training set for salient object detection.
For a fair comparison, we adopted ResNet-101 as the backbone network for all the methods.
%
Table~\ref{table:non-local} reports the results, where our method still achieves the best performance on all the benchmark datasets, which proves the effectiveness of the designed SAC module.

\begin{figure} [tp]
	\centering
	\includegraphics[width=0.99\linewidth]{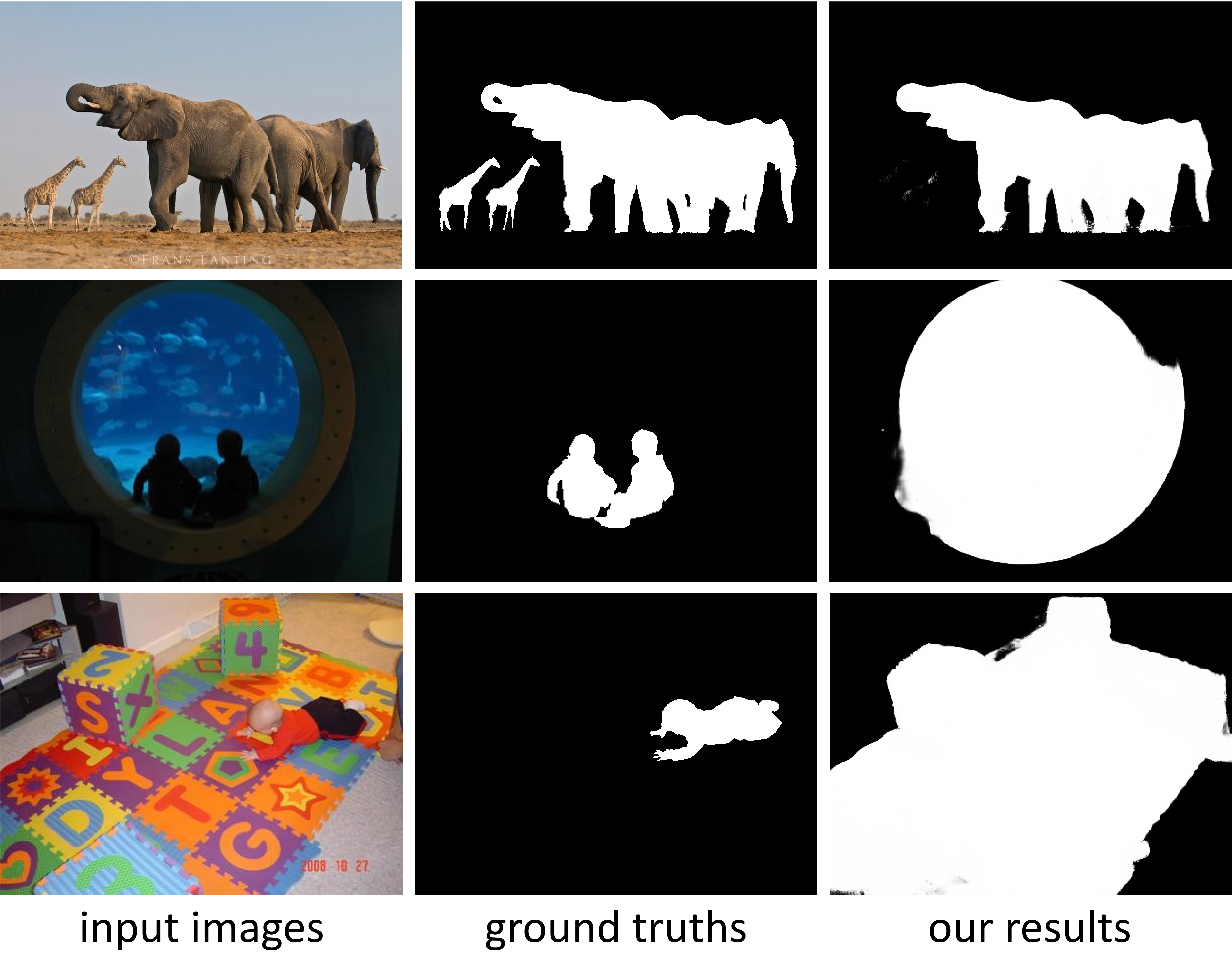}
	\caption{Three typical failure cases.}
	\label{fig:fail_case}
\end{figure}

Lastly, we also analyzed the failure cases, for which we found to be highly challenging.
%
For instance, our method may fail for
(i) multiple salient objects in very different scales (see Fig.~\ref{fig:fail_case} (top)), where the network may regard the small objects as non-salient background;
(ii) dark salient objects (see Fig.~\ref{fig:fail_case} (middle)), where there are insufficient context to determine whether the regions are salient or not; and
(iii) salient objects over a complex background (see Fig.~\ref{fig:fail_case} (bottom)), where high-level scene knowledge is required to understand the image.



\section{Conclusion}
\label{sec:conclusion}

This paper presents a novel saliency detection network based on the spatial attenuation context.  
Our key idea is to recurrently propagate and aggregate image context with different attenuation factors and to integrate the aggregated features using weights learnt from an attention mechanism.
Using our model, local image context can adaptively propagate over different ranges, and 
we can leverage the complementary advantages of these context to improve the saliency detection quality.
In the end, we evaluated our method on six common benchmark datasets and compared it extensively with 29 state-of-the-art methods.
Experimental results clearly show that our method performs favorably over all the others, both visually and quantitatively.
In the future, we plan to explore the potential of our SAC module design for instance-level salient object detection and enhance its capability for detecting salient objects in videos.

	%

	{\small
		\bibliographystyle{IEEEtran}
		\bibliography{reference}
	}


	\ifCLASSOPTIONcaptionsoff
	\newpage
	\fi

	\begin{IEEEbiography}[{\includegraphics[width=1in,height=1.25in,clip,keepaspectratio]{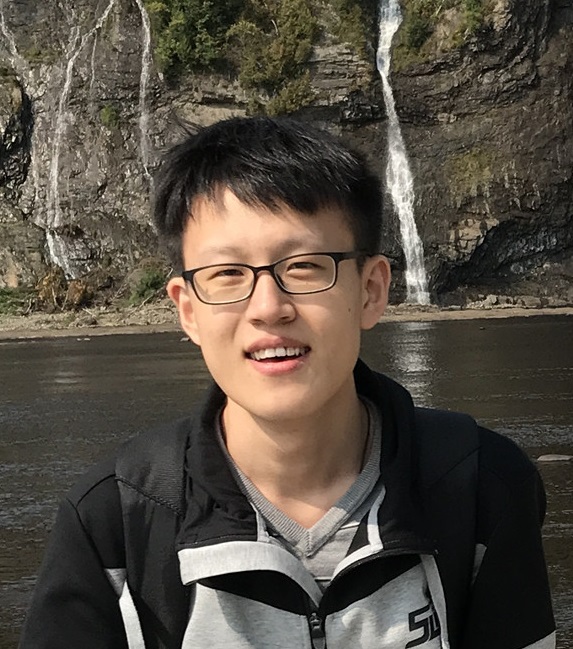}}]{Xiaowei Hu}
		
		received his B.Eng. degree in Computer Science and Technology from South China University of Technology, China, in 2016. He is currently working toward the Ph.D. degree with the Department of Computer Science and Engineering, The Chinese University of Hong Kong. His research interests include computer vision, deep learning, and low-level vision.
		
	\end{IEEEbiography}

	\vspace{-5mm}
	\begin{IEEEbiography}[{\includegraphics[width=1.0in,height=1.25in,clip,keepaspectratio]{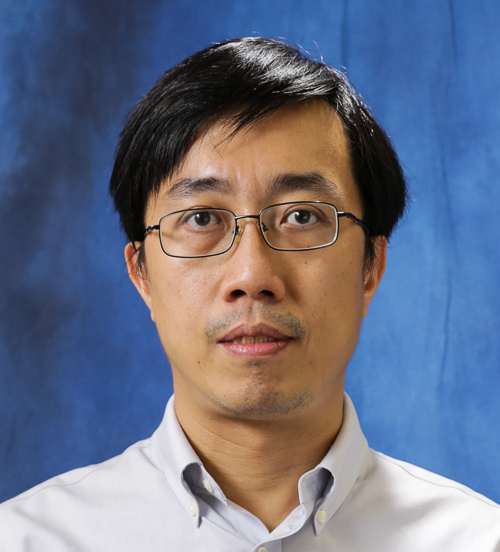}}]{Chi-Wing Fu} is currently an associate professor in the Chinese University of Hong Kong.  He served as the co-chair of SIGGRAPH ASIA 2016's Technical Brief and Poster program, associate editor of Computer Graphics Forum, and panel member in SIGGRAPH 2019 Doctoral Consortium, as well as program committee members in various research conferences, including SIGGRAPH Asia Technical Brief, SIGGRAPH Asia Emerging tech., IEEE visualization, CVPR, IEEE VR, VRST, Pacific Graphics, GMP, etc.  His recent research interests include computation fabrication, 3D computer vision, user interaction, and data visualization.
	\end{IEEEbiography}

	\vspace{-5mm}
	\begin{IEEEbiography}[{\includegraphics[width=1in,height=1.25in,clip,keepaspectratio]{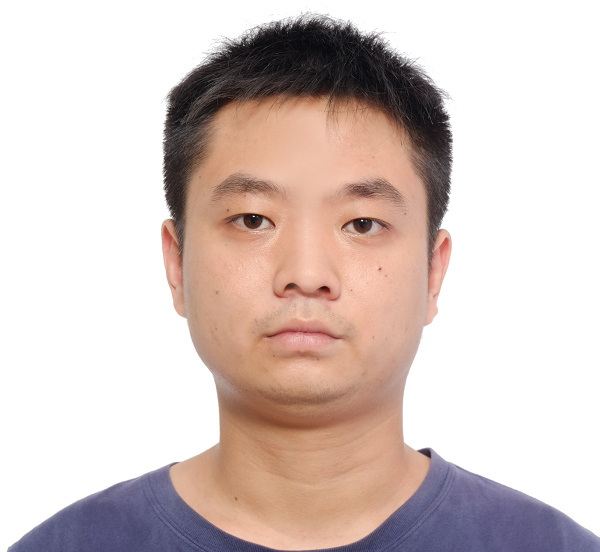}}]{Lei Zhu}
		received his Ph.D. degree in the Department of Computer Science and Engineering from the Chinese University of Hong Kong in 2017.
		He is working as a postdoctoral fellow at the Chinese University of Hong Kong. His research interests include computer graphics, computer vision, medical image processing, and deep learning.
	\end{IEEEbiography}
	
	
	\vspace{-5mm}
	\begin{IEEEbiography}[{\includegraphics[width=1in,height=1.25in,clip,keepaspectratio]{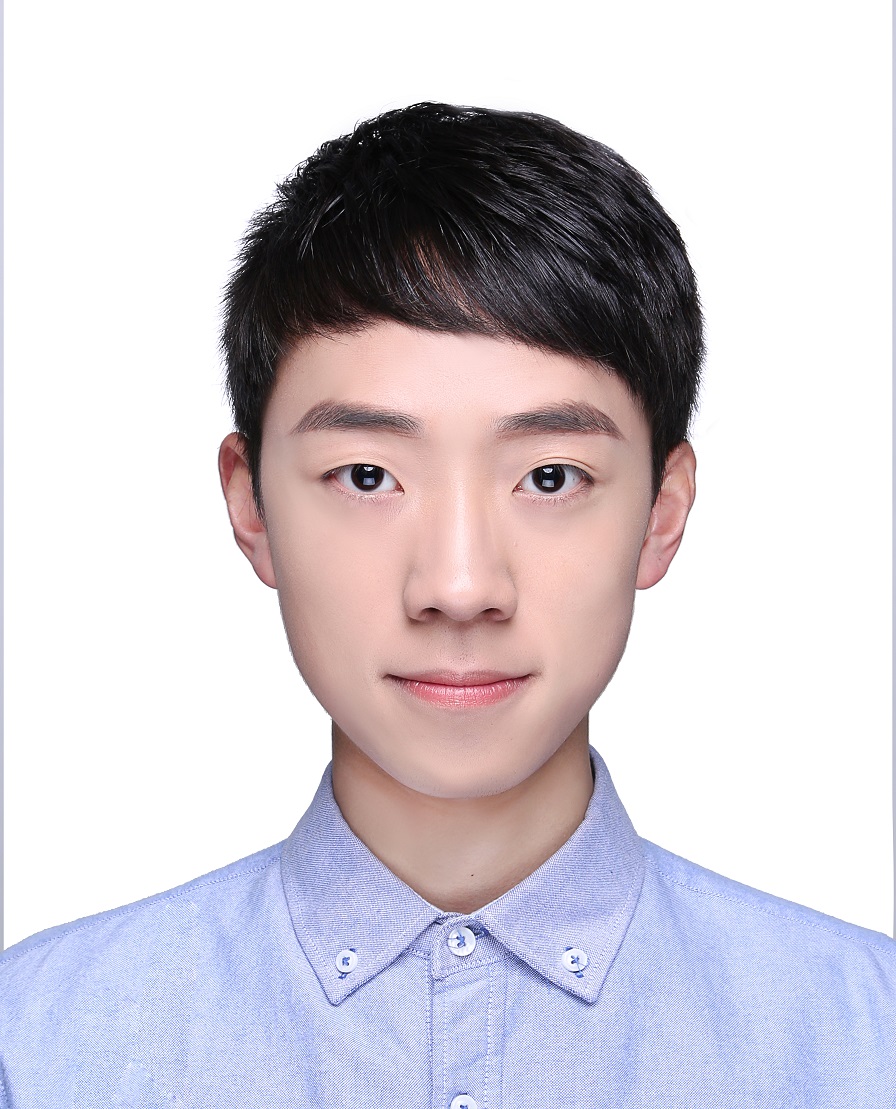}}]{Tianyu Wang} received his B.Eng. degree in Computer Science and Technology from Dalian University of Technology, China, in 2018. He is currently working as a research assistant at the Chinese University of Hong Kong. His research interests include computer vision, image processing, computational photography, low-level vision, and deep learning.
	\end{IEEEbiography}
	
	\vspace{-5mm}
	\begin{IEEEbiography}[{\includegraphics[width=1in,height=1.25in,clip,keepaspectratio]{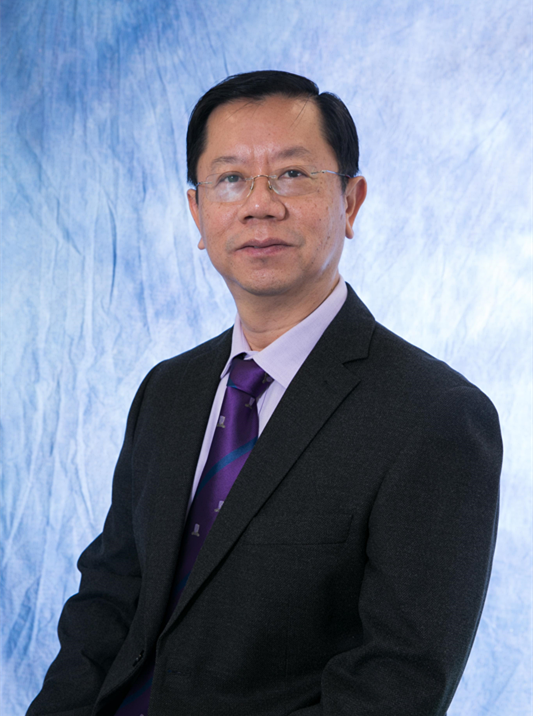}}]{Pheng-Ann Heng} received his B.Sc. (Computer Science) from the National University of Singapore in 1985. 
		He received his M.Sc. (Computer Science), M. Art (Applied Math) and Ph.D. (Computer Science) all from the Indiana University in 1987, 1988, 1992 respectively.
		He is a professor at the Department of Computer Science and Engineering at The Chinese University of Hong Kong. He has served as the Department Chairman from 2014 to 2017 and as the Head of Graduate Division from 2005 to 2008 and then again from 2011 to 2016.
		He has served as the Director of Virtual Reality, Visualization and Imaging Research Center at CUHK since 1999. He has served as the Director of Center for Human-Computer Interaction at Shenzhen Institutes of Advanced Technology, Chinese Academy of Sciences since 2006. He has been appointed by China Ministry of Education as a Cheung Kong Scholar Chair Professor in 2007. 
		His research interests include AI and VR for medical applications, surgical simulation, visualization, graphics, and human-computer interaction.
	\end{IEEEbiography}

\end{document}